\documentclass[11pt,a4paper]{article}
\usepackage[utf8]{inputenc}
\usepackage{array}
\usepackage{booktabs}
\usepackage{url}
\usepackage{amstext}
\usepackage{graphicx}

\makeatletter

\pdfpageheight\paperheight
\pdfpagewidth\paperwidth

\DeclareTextSymbolDefault{\textquotedbl}{T1}
\providecommand{\tabularnewline}{\\}

\usepackage{placeins}
\usepackage{springdrift-common}

\title{Springdrift: An Auditable Persistent Runtime for LLM Agents\\with Case-Based Memory, Normative Safety, and Ambient Self-Perception}
\author{Seamus Brady}
\date{March 2026}

\makeatother

\usepackage{listings}

\begin{document}
\begin{center}
\includegraphics[width=0.35\textwidth]{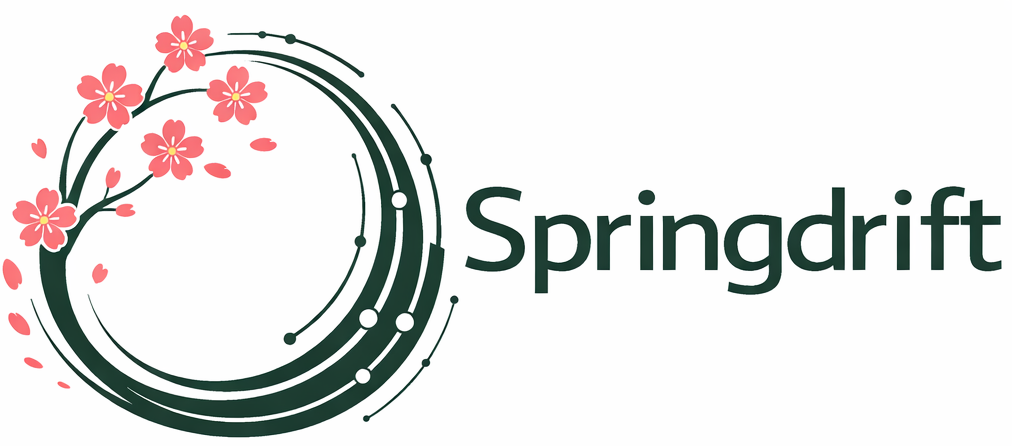} 
\par\end{center}

\vspace{-0.5em}

\maketitle
\begin{abstract}
We present Springdrift, a persistent runtime for long-lived LLM agents.
The system integrates an auditable execution substrate (append-only
memory, supervised processes, git-backed recovery), a case-based reasoning
memory layer with hybrid retrieval (evaluated on a synthetic benchmark
against a dense cosine baseline), a deterministic normative calculus
for normative resolution in safety gating, with auditable axiom trails, and continuous ambient
self-perception via a structured self-state representation (the \emph{sensorium})
injected each cycle without tool calls.

The integration of these properties supports behaviours that are difficult
to achieve in session-bounded systems: cross-session task continuity,
cross-channel context maintenance, end-to-end forensic reconstruction
of any decision, and self-diagnostic behaviour observed during deployment.
We report on a single-instance, single-operator deployment over a 23-day
period (19~operating days), during which the agent diagnosed its own
infrastructure bugs, classified its own failure modes, identified an
architectural vulnerability, and maintained context across email and web
channels---without explicit instruction.

We introduce the term \emph{Artificial Retainer} for this category
of system: a non-human system with persistent memory, defined authority,
domain-specific autonomy, and forensic accountability, engaged in
an ongoing relationship with a specific principal --- a category
we distinguish from both software assistants and autonomous agents,
drawing instead on the engagement structure of professional retainer
relationships and the bounded autonomy of trained working animals.

This is a technical report describing a systems design and deployment
case study, not a benchmark-driven evaluation. The evidence is drawn
from a single instance with a single operator; we present it as an illustration
of what these architectural properties can support in practice, not
as proof of general effectiveness.

The system is implemented in approximately 62,000~lines of Gleam
on Erlang/OTP, with approximately 1,500~tests. All code, evaluation artefacts,
and redacted operational logs (494~narrative entries, 24,035~cycle
log entries) will be available at \url{https://github.com/seamus-brady/springdrift} upon publication. 
\end{abstract}

\section{Introduction}


Many current LLM-based agents are effectively session-bounded systems.
A conversation begins, tools are called, a response is produced, and
the local context is discarded or heavily compressed. This is often
sufficient for short interactions. It is much less satisfactory for
agents expected to operate over weeks or months, accumulate experience,
maintain consistent conduct, and remain inspectable to their operators.

Long-lived operation changes what matters. An operator who works with
an agent over extended periods needs more than competent one-off responses.
They need the agent's conduct to be predictable across sessions, its
refusals to be explainable, its failures to become visible rather
than remaining silent, and its decisions to be reconstructable after
the fact. These are not only usability concerns. They are preconditions
for trust.

\textbf{Design thesis.} Long-lived agents should be built as
auditable, self-observing operational systems with stable normative
commitments --- not as stateless tools that happen to persist.

Current agent frameworks optimise for task completion within a
session. They do not provide the invariants required for long-lived,
inspectable operation: forensic reconstruction of past decisions,
persistent self-observation, or structured authority that the operator
can audit and adjust. Adding memory to a session-bounded system does
not solve this --- it adds recall without accountability, persistence
without auditability.

This paper presents Springdrift, a persistent runtime for long-lived
LLM agents. Rather than a framework for constructing agents, Springdrift
is an execution environment that provides process supervision, append-only
memory, structured safety evaluation, continuous self-perception,
and full operational auditability. An agent in Springdrift is not
re-instantiated per session but runs continuously as a supervised
process with durable state and replayable history.

We present Springdrift not as a finished product but as a
\emph{reference architecture}: a concrete instantiation of a design
space that we believe is underexplored. The architecture described in
this paper is, in retrospect, a reference implementation of what we
term the \emph{Artificial Retainer} --- a concept we define in
Section~\ref{sec:artificial-retainer} and distinguish from existing
categories of AI system. Individual components (CBR, normative
calculus, sensorium) are separable; the contribution is the
integration and the design thesis it embodies. Readers should
distinguish the \emph{core invariants} (auditability, persistence,
self-observation) from the \emph{implementation choices} (Gleam/OTP,
XStructor, Stoic normative framework). The invariants are the thesis;
the choices are one way to realise it.

\subsection{Contributions}

We describe four integrated contributions, ordered by what we
consider their importance to the design thesis:
\begin{enumerate}
\item \textbf{Full operational auditability} (Section~\ref{sec:runtime})
--- append-only memory stores, cycle-level decision logging, and
git-backed state recovery that make every agent decision forensically
reconstructable after the fact. We argue this is a \emph{prerequisite}
for trust in long-lived agents, not merely a debugging convenience:
an agent whose decisions cannot be inspected cannot be trusted,
regardless of how capable it is.
\item \textbf{Continuous ambient self-perception} via the sensorium (Section~\ref{sec:sensorium}),
a structured self-state block injected into the agent's context each
cycle without tool calls, giving the agent persistent awareness of
its own operational state.
\item \textbf{A case-based reasoning memory layer} (Section~\ref{sec:cbr})
with hybrid lexical--semantic retrieval, evaluated against a dense
cosine baseline on a controlled benchmark of 800~cases and 200~queries.
\item \textbf{A deterministic normative calculus} (Section~\ref{sec:normative})
inspired by Becker's Stoic ethics, producing auditable axiom trails
for every safety decision, evaluated for completeness, determinism,
and monotonicity over the full proposition space.
\end{enumerate}
We also present a design argument (Section~\ref{sec:character})
that long-lived agents benefit from stable character (persistent normative
commitments) and self-observation capacity, and illustrate this with
deployment observations.

These contributions occupy different levels of necessity. We argue that
\emph{auditability} (append-only logs, forensic replay, decision
reconstruction) is a prerequisite for any long-lived agent that needs
to be trusted --- without it, the operator cannot verify what happened
or why. \emph{Persistent memory} and \emph{self-perception} are
strongly beneficial, enabling cross-session continuity and proactive
self-monitoring. The \emph{normative calculus} and \emph{case-based
reasoning} are experimental design bets --- architecturally motivated
but not yet validated beyond a single deployment. We encourage readers
to reason about adoption at these different levels rather than treating
the system as all-or-nothing.

\subsection{Scope}

This paper is a systems and deployment report: its purpose is to
document an architecture, its invariants, and the behaviours it
enabled in practice, rather than to establish comparative benchmark
superiority. The goal is not to optimise benchmark performance, but
to explore the design space of long-lived, inspectable agent systems.
We treat the system as an engineering artefact and report its observed
behaviour under deployment. The evidence is drawn from a single instance
($n=1$) and is anecdotal; its role is illustrative, not conclusive.

We are not claiming general performance improvements over existing
agent frameworks. We are not claiming safety guarantees --- the
normative calculus is a structured decision procedure, not a proof
system. We are not claiming generality across domains or operators.
What we claim is that the \emph{combination} of architectural
properties described here enables behaviours that are difficult to
achieve without them, and that the design space is worth exploring.

The design draws on Sloman's H-CogAff cognitive architecture \citep{sloman2001,sloman2003},
Becker's \emph{A New Stoicism} \citep{becker1998}, Beach's work on
narrative thought and decision making \citep{beach2010}, Packer et
al.'s MemGPT \citep{packer2023}, and the System~M framework of \citet{dupoux2026}.


\section{Runtime Architecture}

\label{sec:runtime} 

\subsection{Overview}

Springdrift runs as an Erlang/OTP application. The cognitive loop
is the central process---an OTP actor that receives inputs (from
the operator via terminal or web GUI, from the scheduler, or from
the email inbox poller), orchestrates specialist sub-agents, manages
safety evaluation, and produces outputs. All components are supervised
OTP processes communicating via typed \texttt{Subject(T)} channels.
No shared mutable state.

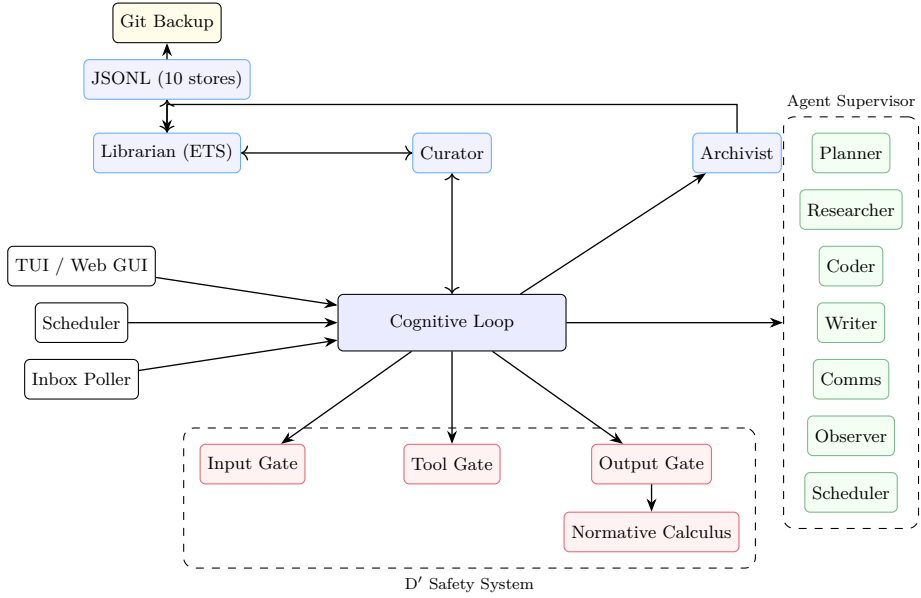
\begin{figure}[ht]
\centering \begin{tikzpicture}[scale=0.75, transform shape,
  every node/.style={font={\footnotesize}},
  box/.style={draw, rounded corners=2pt, minimum height=0.7cm, align=center, fill=white, line width=0.4pt},
  mem/.style={box, fill=blue!5, draw=sdblue!60},
  act/.style={box, fill=sdlightbg},
  agt/.style={box, fill=green!5, draw=sdgreen!60},
  saf/.style={box, fill=red!5, draw=sdred!60},
]
  \node[mem] (librarian) at (-5, 5) {Librarian (ETS)};
  \node[mem] (curator) at (0, 5) {Curator};
  \node[mem] (archivist) at (5, 5) {Archivist};
  \node[mem] (jsonl) at (-5, 6.3) {JSONL (10 stores)};
  \node[box, fill=yellow!10] (git) at (-5, 7.3) {Git Backup};

  \node[act, minimum width=4cm, minimum height=1cm, fill=blue!8] (cognitive) at (0, 2) {Cognitive Loop};

  \node[box] (tui) at (-6.5, 3) {TUI / Web GUI};
  \node[box] (scheduler) at (-6.5, 2) {Scheduler};
  \node[box] (poller) at (-6.5, 1) {Inbox Poller};

  \node[agt] (planner) at (7, 5) {Planner};
  \node[agt] (researcher) at (7, 4) {Researcher};
  \node[agt] (coder) at (7, 3) {Coder};
  \node[agt] (writer) at (7, 2) {Writer};
  \node[agt] (comms) at (7, 1) {Comms};
  \node[agt] (observer) at (7, 0) {Observer};
  \node[agt] (schedagent) at (7, -1) {Scheduler};
  \node[draw, dashed, rounded corners,
        fit=(planner)(researcher)(coder)(writer)(comms)(observer)(schedagent),
        inner sep=6pt, label={[font={\scriptsize}]above:Agent Supervisor}] (sup) {};

  \node[saf] (input_gate) at (-3.5, -0.5) {Input Gate};
  \node[saf] (tool_gate) at (0, -0.5) {Tool Gate};
  \node[saf] (output_gate) at (3.5, -0.5) {Output Gate};
  \node[saf] (normcalc) at (3.5, -1.7) {Normative Calculus};
  \node[draw, dashed, rounded corners,
        fit=(input_gate)(tool_gate)(output_gate)(normcalc),
        inner sep=6pt, label={[font={\scriptsize}]below:D$'$ Safety System}] {};

  \draw[sdarrow] (tui) -- (cognitive);
  \draw[sdarrow] (scheduler) -- (cognitive);
  \draw[sdarrow] (poller) -- (cognitive);

  \draw[sdarrow] (cognitive) -- (sup);

  \draw[sdarrow, <->] (cognitive) -- (curator);
  \draw[sdarrow, <->] (librarian) -- (curator);
  \draw[sdarrow] ([xshift=1.2cm]cognitive.north) -- (archivist);
  \draw[sdarrow] (archivist.north) -- ++(0,0.5) -| (librarian.north);
  \draw[sdarrow, <->] (librarian) -- (jsonl);
  \draw[sdarrow] (jsonl) -- (git);

  \draw[sdarrow] (cognitive) -- (input_gate);
  \draw[sdarrow] (cognitive) -- (tool_gate);
  \draw[sdarrow] (cognitive) -- (output_gate);
  \draw[sdarrow] (output_gate) -- (normcalc);
\end{tikzpicture} \caption{System architecture. The cognitive loop orchestrates all components
via typed message-passing channels. This is a simplified diagram; the system can also create teams of cooperating subagents.}
\label{fig:architecture} 
\end{figure}

\subsection{Configuration and Continuity}

Each instance maintains: a stable UUID (generated on first run, never
changed), a persona document, a character specification of formal
normative propositions, and a session preamble template assembled
from live memory state each cycle.

\subsection{Memory Subsystem}

Ten append-only JSONL stores are indexed in ETS by the Librarian
actor. Current state is derived by replaying operations chronologically---no
record is modified or deleted during normal operation.

\begin{table}[ht]
\centering{\small{}%
\begin{tabular}{lll}
\toprule 
{\small Store } & {\small Record Type } & {\small Purpose }\tabularnewline
\midrule 
{\small Narrative } & {\small NarrativeEntry } & {\small What happened each cycle }\tabularnewline
{\small Threads } & {\small Thread } & {\small Topics grouping related entries }\tabularnewline
{\small Facts } & {\small MemoryFact } & {\small Key-value memory (scoped, decayed) }\tabularnewline
{\small CBR cases } & {\small CbrCase } & {\small Problem-solution-outcome with utility }\tabularnewline
{\small Artifacts } & {\small ArtifactRecord } & {\small Large content on disk }\tabularnewline
{\small Tasks } & {\small PlannerTask } & {\small Planned work with steps }\tabularnewline
{\small Endeavours } & {\small Endeavour } & {\small Multi-task goals }\tabularnewline
{\small Comms } & {\small CommsMessage } & {\small Sent and received email }\tabularnewline
{\small Affect } & {\small AffectSnapshot } & {\small Functional emotion readings }\tabularnewline
{\small DAG nodes } & {\small CycleNode } & {\small Per-cycle telemetry }\tabularnewline
\bottomrule
\end{tabular}}{\small\caption{The ten memory stores.}
}
\end{table}

The \textbf{Curator} (\texttt{narrative/curator.gleam}) assembles
a virtual context window each cycle, following the Letta-style approach
of treating the LLM's context as a managed resource rather than a
passive conversation history. The context window is divided into priority-budgeted
slots, each assigned a priority level from~1 (identity) through~10
(background). At assembly time, the Curator loads the agent's persona
and character specification, queries the Librarian for active thread
count, persistent fact count, and CBR case count, renders the sensorium
XML (Section~\ref{sec:sensorium}), retrieves relevant CBR cases
organised by category, and populates the session preamble template
using \texttt{\{\{slot\}\}} substitution with \texttt{{[}OMIT IF{]}}
conditional rules. When total content exceeds the configurable character
budget (\texttt{preamble\_budget\_chars}, default 8000~characters,
approximately 2000~tokens), lower-priority slots are truncated or
cleared---identity and sensorium are never shed. The result is that
the agent perceives a coherent, current view of its own state without
making any tool calls.

The \textbf{Archivist} runs asynchronously after each cycle as a fire-and-forget
process, using a two-phase pipeline based on the ACE Reflector/Curator
architecture \citep{ace2024}: Phase~1 generates an honest plain-text
reflection of what worked and what failed; Phase~2 takes that reflection
and produces structured \texttt{NarrativeEntry} and \texttt{CbrCase}
records via XStructor XML schema validation. If Phase~1 fails, a
single-call fallback is used. If Phase~2 fails after Phase~1 succeeds,
the reflection is preserved in logs. Thread assignment uses overlap
scoring across location, domain, and keyword dimensions. CBR cases
track their own utility via retrieval and outcome statistics, enabling
self-improving retrieval over time. Detailed descriptions of XStructor
and all other technical subsystems (canary probes, cross-cycle pattern
detection, the agent-managed scheduler, narrative threading, housekeeping,
the sandbox, and the web GUI) are in Appendix~\ref{app:subsystems}.

\subsection{Cognitive Cycle}

Each cognitive cycle follows a fixed sequence: input reception, D$'$
input gate evaluation, query complexity classification (routing to
faster or more capable models), context assembly by the Curator (including
sensorium injection), LLM inference, tool dispatch with per-tool safety
evaluation, react loop until text response, output gate evaluation,
response delivery, and asynchronous Archivist processing.

The \textbf{D$'$ safety system} (detailed in Section~\ref{sec:normative})
evaluates at three points in this sequence. The \emph{input gate}
screens incoming prompts for injection attempts using canary probes
and deterministic regex rules, with a fast-accept path that skips
LLM scoring for benign operator input. The \emph{tool gate} evaluates
each tool call before dispatch---this gate always runs the full LLM
scorer, since the threat model here is a compromised agent acting
on injected instructions from web content. The \emph{output gate}
evaluates finished responses before delivery, using different strategies
for interactive cycles (deterministic rules only---the operator is
the quality gate) and autonomous cycles (full LLM scorer plus normative
calculus). The name D$'$ comes from signal detection theory: the
system attempts to maximise discriminability between legitimate and
unsafe operations.

An important design decision is the \textbf{interactive/autonomous
split}. Interactive input (operator typing) skips structural injection
heuristics---the operator may legitimately discuss the system's own
safety mechanisms. Autonomous input (scheduler, email) uses the full
evaluation stack.

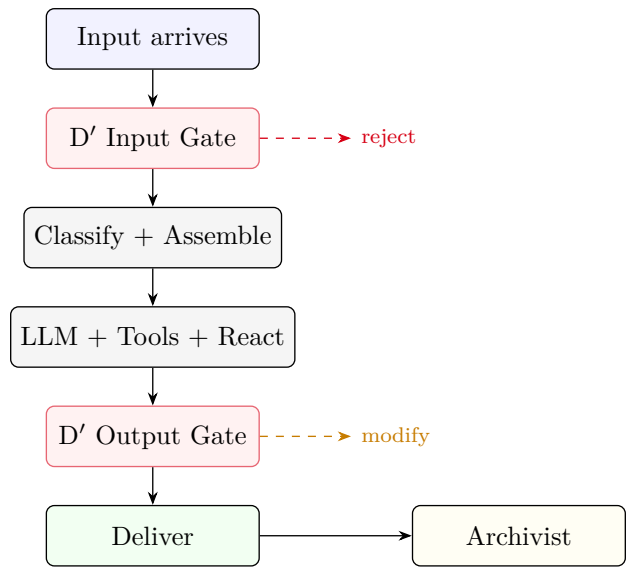
\begin{figure}[ht]
\centering \begin{tikzpicture}[node distance=0.5cm, every node/.style={font={\small}}]
  \node[sdbox, fill=blue!5] (input) {Input arrives};
  \node[sdsafety, below=of input] (igate) {D$'$ Input Gate};
  \node[sdactor, below=of igate] (classify) {Classify + Assemble};
  \node[sdactor, below=of classify] (llm) {LLM + Tools + React};
  \node[sdsafety, below=of llm] (ogate) {D$'$ Output Gate};
  \node[sdbox, fill=green!5, below=of ogate] (deliver) {Deliver};
  \node[sdbox, fill=yellow!5, right=2cm of deliver] (arch) {Archivist};
  \draw[sdarrow] (input) -- (igate);
  \draw[sdarrow] (igate) -- (classify);
  \draw[sdarrow] (classify) -- (llm);
  \draw[sdarrow] (llm) -- (ogate);
  \draw[sdarrow] (ogate) -- (deliver);
  \draw[sdarrow] (deliver) -- (arch);
  \draw[sdarrow, sdred, dashed] (igate.east) -- ++(1.2,0) node[right, font={\scriptsize}, text=sdred] {reject};
  \draw[sdarrow, sdorange, dashed] (ogate.east) -- ++(1.2,0) node[right, font={\scriptsize}, text=sdorange] {modify};
\end{tikzpicture} \caption{Cognitive cycle. Every input passes through safety gates before and
after inference.}
\label{fig:cycle} 
\end{figure}

\subsection{Specialist Agents}

The cognitive loop delegates work to specialist agents rather than
attempting everything in a single react loop. Each agent is a supervised
OTP process with its own message history, tool set, context window,
and react loop. The agent framework (\texttt{agent/framework.gleam})
wraps each \texttt{AgentSpec} into a running process: it calls the
LLM, executes tool calls, and loops until it receives a text response
or hits its turn limit.

\begin{table}[ht]
\centering{\small{}%
\begin{tabular}{llcll}
\toprule 
{\small Agent } & {\small Tools } & {\small Turns } & {\small Restart } & {\small Role }\tabularnewline
\midrule 
{\small Planner } & {\small none (XML output) } & {\small 5 } & {\small Permanent } & {\small Plan decomposition, risk identification }\tabularnewline
{\small Project Manager } & {\small 22 planner tools } & {\small 8 } & {\small Permanent } & {\small Tasks, endeavours, phases, blockers }\tabularnewline
{\small Researcher } & {\small web + artifacts } & {\small 8 } & {\small Permanent } & {\small Search, extraction, source tracking }\tabularnewline
{\small Coder } & {\small sandbox (6 tools) } & {\small 10 } & {\small Permanent } & {\small Code execution in Podman containers }\tabularnewline
{\small Writer } & {\small built-in } & {\small 6 } & {\small Permanent } & {\small Text drafting and editing }\tabularnewline
{\small Comms } & {\small 4 email tools } & {\small 6 } & {\small Permanent } & {\small Email via AgentMail }\tabularnewline
{\small Observer } & {\small 18 diagnostic tools } & {\small 6 } & {\small Transient } & {\small Cycle forensics, CBR curation }\tabularnewline
\bottomrule
\end{tabular}}{\small\caption{Specialist agents. Turn limits are defaults, configurable per profile.}
\label{tab:agents} }
\end{table}

Three coordination mechanisms are available. \textbf{Direct delegation}:
the cognitive loop dispatches a single agent via a tool call. \textbf{Parallel
dispatch}: when the LLM requests multiple agent tool calls in one
response, all agents run simultaneously as independent OTP processes;
results are accumulated in any order and combined for synthesis. \textbf{Agent
teams}: coordinated groups with a shared strategy---\texttt{ParallelMerge}
(all work simultaneously, results merged by synthesis LLM), \texttt{TeamPipeline}
(sequential, each receives prior output), \texttt{DebateAndConsensus}
(independent analyses then debate over rounds with convergence detection),
and \texttt{LeadWithSpecialists} (specialists work in parallel, then
a lead synthesises). Teams appear as tools (prefixed \texttt{team\_})
alongside individual agent tools.

When an agent completes, the framework populates typed \texttt{AgentFindings}
(e.g.\ \texttt{ResearcherFindings} with sources and dead ends, \texttt{CoderFindings}
with files touched). Tool errors that occurred during the react loop
are captured in \texttt{AgentSuccess.tool\_errors}---when non-empty,
the cognitive loop prefixes a warning so the orchestrating LLM knows
the result may be unreliable. Agent health is also pushed to the Curator's
sensorium via \texttt{UpdateAgentHealth}, making degradation visible
before the next cycle.

Delegation depth is capped by \texttt{max\_delegation\_depth} (default~3).
Sub-agents cannot call \texttt{request\_human\_input}---they report
only through their return value, preventing a delegated agent from
hijacking the operator interaction channel. Active delegations are
tracked on \texttt{CognitiveState} and rendered in the sensorium's
\texttt{\textless delegations\textgreater} section, showing turn
progress, token usage, and elapsed time.

The \textbf{comms agent} sends and receives email with three independent
safety layers: a hard recipient allowlist (not D$'$-bypassable),
deterministic regex rules blocking credentials and system internals,
and agent-specific D$'$ features with tighter tool gate thresholds
(0.30/0.50 vs the default 0.35/0.55). An inbox polling actor checks
for new messages at a configurable interval and routes them to the
cognitive loop as \texttt{SchedulerInput}, triggering the full autonomous
evaluation stack.

\subsection{Auditability and Git Backup}

All runtime state is reconstructable from append-only JSONL files.
No record is modified or deleted during normal operation; current
state is derived by replaying operations chronologically. This makes
forensic reconstruction straightforward: any fact, any case, any task
can be traced to the cycle that created it.

The \textbf{cycle log} (daily-rotated JSONL in \texttt{.springdrift/memory/cycle-log/})
records every LLM call, tool execution, D$'$ gate decision, and agent
delegation as structured DAG nodes. Nodes form a parent-child tree:
a user or scheduler cycle at the root, with agent sub-cycles and tool
calls as descendants. Each node carries token counts, model used,
elapsed time, and typed agent output. The Observer agent's \texttt{inspect\_cycle}
and \texttt{list\_recent\_cycles} tools provide the agent itself with
drill-down access to this telemetry.

The \textbf{Librarian} actor owns ETS tables that serve as a fast
query cache over the JSONL stores. At startup, it replays up to \texttt{librarian\_max\_days}
(default~30) of narrative entries, CBR cases, facts, artifacts, and
DAG nodes into in-memory tables. All memory tools query through the
Librarian when available, falling back to direct JSONL reads when
it is not running. New entries written by the Archivist or other subsystems
are indexed immediately via notification messages, so the cache is
always current within the session.

\textbf{Housekeeping} (\texttt{narrative/housekeeping.gleam}) runs
periodically to maintain memory quality. Three operations: CBR deduplication
(symmetric weighted field similarity above a configurable threshold
merges cases), case pruning (old low-confidence failures without pitfall
annotations are removed), and fact conflict resolution (when multiple
facts share a key, the higher-confidence value is retained). Fact
confidence decays at read time via a half-life formula ($c_{t}=c_{0}\cdot2^{-t/h}$,
default half-life 30~days), so stale facts naturally lose influence
without requiring explicit cleanup.

The \texttt{.springdrift/} directory is a git repository with periodic
commits and optional remote push. A standalone Python tool (SD~Audit)
provides a web dashboard with Chart.js visualisations for offline
analysis of daily activity, token usage, D$'$ gate decisions, and
outcome distributions.

\subsection{Implementation}

Springdrift is implemented in Gleam \citep{gleam2022}, a statically
typed language that compiles to Erlang bytecode and runs on the BEAM
virtual machine. This choice is pragmatic, not aesthetic---the BEAM
and Gleam's type system provide specific properties that matter for
long-lived agent systems. A detailed technical overview is in Appendix~\ref{app:beam}.

\textbf{Why not Python?} Every major agent framework (LangGraph, CrewAI,
AutoGen) is Python. Python's strengths (ecosystem, LLM library support)
are real, but for a persistent runtime that manages concurrent processes,
handles failures during multi-step reasoning, and must remain auditable
over weeks of operation, its weaknesses are structural: the GIL limits
true concurrency, exception handling is advisory rather than enforced,
and there is no supervision model for automatic failure recovery.

\textbf{BEAM runtime properties.} The BEAM was designed for telecommunications
systems that run continuously and handle failures gracefully. Each
Springdrift component---cognitive loop, Librarian, Curator, Archivist,
scheduler, inbox poller, sandbox manager, each specialist agent---is
a supervised OTP process. If a process crashes, its supervisor restarts
it according to a configured strategy (Permanent, Transient, Temporary).
The system continues operating. In deployment, we observed this directly: during the
23-day deployment, 714~LLM timeouts and multiple sub-agent crashes
were handled by supervision without manual intervention (Section~\ref{sec:runtime-chars}).
Preemptive scheduling means one stuck LLM call does not block the
scheduler or inbox poller. Per-process garbage collection means no
stop-the-world pauses during long-running operation.

\textbf{Type-system guarantees.} Gleam's static types catch
malformed tool call parameters, missing message handler variants,
and protocol mismatches at compile time. The \texttt{Result} type
forces every error path to be handled explicitly---there are no unchecked
exceptions. For agent systems where a single unhandled error can derail
a multi-step reasoning chain, this compounds: every tool call, every
message send, every JSON decode is a \texttt{Result} that the compiler
requires be addressed.

\textbf{State ownership.} All cross-process communication uses typed
\texttt{Subject(T)} channels. No shared mutable state, no locks, no
race conditions on memory access. The Librarian owns all ETS tables.
The Curator owns the virtual context window. The scheduler owns the
job state. Boundaries are enforced by the process model, not by convention.
This is directly relevant to auditability: you can reason about state
ownership because the architecture enforces it.

The system compiles to approximately 62,000~lines across 125~source
files, with approximately 1,500~tests covering memory round-trips, safety gate
behaviour, normative calculus completeness, tool definitions, and
protocol encoding.


\section{The Sensorium: Continuous Self-Perception}

\label{sec:sensorium} 

\subsection{Motivation}

Between tool calls, most agent systems are perceptually blind. The
agent does not know what time it is, how its recent work has gone,
or whether its subsystems are healthy. Each diagnostic query costs
tokens and turns---a bootstrapping problem where the agent must decide
to check before it has the information that would tell it whether
checking is worthwhile.

\subsection{Design}

The sensorium is a structured XML block injected into the system prompt
at every cycle. It is computed by the Curator from live system state
and persistent narrative history. No tool calls, no LLM queries. It
contains: Clock (time, uptime, cycle ID), Situation (input source,
queue depth, active thread), Schedule (pending/overdue jobs), Vitals
(rolling performance summary), Delegations (active sub-agent status),
and Tasks (planned work).

\clearpage\bgroup
\begin{lstlisting}[language=XML,caption={Sensorium example (simplified).}]
<sensorium>
  <clock now="2026-03-29T14:30:00"
         session_uptime="2h15m" cycle_id="a7f3b2c1"/>
  <vitals cycles_today="8" agents_active="5"
          success_rate="0.75" cost_trend="stable"
          cbr_hit_rate="0.60" novelty="0.42"
          recent_failures="web_search timeout"/>
</sensorium>
\end{lstlisting}
\leavevmode\egroup

\subsection{Performance Signals}

The vitals carry a rolling performance summary computed from the persistent
narrative log: success rate, cost trend, CBR hit rate, recent failures,
and per-input novelty (keyword Jaccard similarity, following \citealp{dupoux2026}).
These signals span sessions and survive restarts. An earlier version
used session-scoped counters that reset on restart and suffered from
small-sample noise; the history-backed signals eliminated both problems.

\subsection{Operational Consequences of Removing the Sensorium}

The sensorium can be viewed as a structured, zero-latency feature
vector over the agent's operational state, injected into the prompt
each cycle. Without it, the agent starts every cycle with no ambient
context. It does not know the time, cannot calibrate confidence from
recent performance, cannot see overdue jobs, and cannot detect unhealthy
sub-agents. Before the sensorium was implemented, we observed generic
greetings regardless of time, no reference to prior work on session
resume, and delegation decisions without sub-agent health awareness.


\section{Case-Based Retrieval for Agent Memory}

\label{sec:cbr} 

\subsection{Motivation}

Standard dense retrieval does not explicitly represent the structure
of experience or its outcomes. A document retrieved ten times leading
to ten failures is treated identically to one leading to ten successes.
Case-Based Reasoning \citep{aamodt1994} addresses this by representing
experience as structured problem--solution--outcome cases.

\subsection{Retrieval Engine}

Retrieval uses a weighted fusion of six signals: inverted index (0.25),
semantic embedding (0.40), weighted field score (0.10), recency (0.05),
domain match (0.10), and utility score (0.10). Retrieval cap is $K=4$
following \citet{zhou2025}'s context pollution finding.

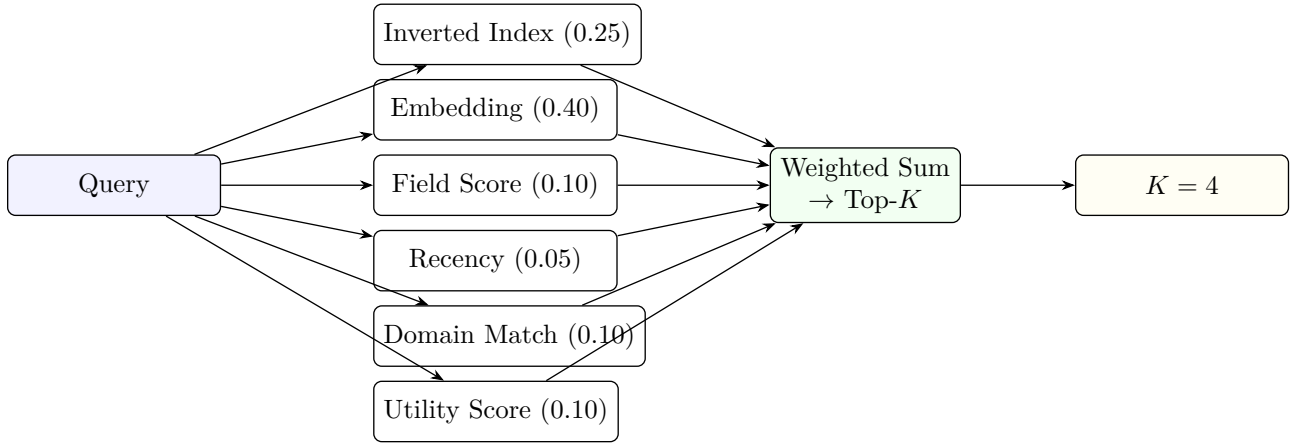
\begin{figure}[ht]
\centering \begin{tikzpicture}[node distance=0.4cm and 1.5cm]
  \node[sdbox, fill=blue!5] (query) {Query};
  \node[sdbox, right=2cm of query, yshift=2cm, minimum width=3.2cm] (index) {Inverted Index (0.25)};
  \node[sdbox, right=2cm of query, yshift=1cm, minimum width=3.2cm] (embed) {Embedding (0.40)};
  \node[sdbox, right=2cm of query, yshift=0cm, minimum width=3.2cm] (field) {Field Score (0.10)};
  \node[sdbox, right=2cm of query, yshift=-1cm, minimum width=3.2cm] (recency) {Recency (0.05)};
  \node[sdbox, right=2cm of query, yshift=-2cm, minimum width=3.2cm] (domain) {Domain Match (0.10)};
  \node[sdbox, right=2cm of query, yshift=-3cm, minimum width=3.2cm] (utility) {Utility Score (0.10)};
  \node[sdbox, right=2cm of field, fill=green!5, minimum width=2.5cm] (fusion) {Weighted Sum\\$\to$ Top-$K$};
  \node[sdbox, right=1.5cm of fusion, fill=yellow!5] (results) {$K=4$};
  \foreach \sig in {index, embed, field, recency, domain, utility} {
    \draw[sdarrow] (query) -- (\sig);
    \draw[sdarrow] (\sig) -- (fusion);
  }
  \draw[sdarrow] (fusion) -- (results);
\end{tikzpicture} \caption{Six-signal retrieval pipeline.}
\label{fig:retrieval} 
\end{figure}

Utility scoring uses Laplace-smoothed outcome tracking: $\text{utility}=(\text{successes}+1)/(\text{retrievals}+2)$.
This is implemented but not evaluated longitudinally in the synthetic
benchmark (all cases have identical usage history).

\subsection{Evaluation}

800~synthetic cases across 4~domains, 200~queries at three difficulty
levels (Easy, Medium, Hard). Ground truth: same domain AND $\geq2$
keyword overlap. Bootstrap 95\% CIs. Notably, the keyword-overlap
component of the ground truth criterion structurally advantages the
lexical retrieval signal; the hybrid advantage on hard queries should
be interpreted with this caution in mind.

\begin{table}[ht]
\centering %
\begin{tabular}{lcccc}
\toprule 
System  & P@4  & 95\% CI  & MRR  & \tabularnewline
\midrule 
Random  & 0.028  & {[}0.018, 0.040{]}  & ---  & \tabularnewline
CBR full (no embed)  & 0.620  & {[}0.574, 0.664{]}  & 0.852  & \tabularnewline
Dense cosine  & 0.920  & {[}0.895, 0.943{]}  & 0.978  & \tabularnewline
CBR index only  & 0.921  & {[}0.897, 0.944{]}  & 0.975  & \tabularnewline
\textbf{CBR hybrid}  & \textbf{0.956}  & \textbf{{[}0.936, 0.974{]}}  & \textbf{0.993}  & \tabularnewline
\bottomrule
\end{tabular}\caption{Retrieval performance ($N=800$, 200~queries, $K=4$). Non-overlapping
CIs for hybrid vs dense cosine.}
\end{table}

On this synthetic benchmark, hybrid CBR outperforms the dense cosine
baseline at every difficulty level, with the advantage most pronounced
on hard queries (0.883 vs 0.796). The win comes from combining lexical index (exact matching
where vocabulary is distinctive) with semantic embedding (cross-vocabulary
similarity where it is not). CBR without embeddings performs significantly
worse (0.620), confirming that the embedding signal is critical.

\begin{figure}[ht]
\centering \begin{tikzpicture}
\begin{axis}[
  width=0.85\textwidth, height=5.5cm,
  xlabel={Case base size}, ylabel={P@4},
  xmin=0, xmax=850, ymin=0.8, ymax=1.02,
  grid=major, grid style={dashed, gray!30},
  legend pos=south east, legend style={font={\small}},
]
\addplot[sdblue, thick, mark=*] coordinates {
  (25, 0.864) (50, 0.850) (100, 0.891) (200, 0.935) (400, 0.948) (800, 0.956)
};
\addlegendentry{Overall}
\addplot[sdred, thick, mark=triangle*, dashed] coordinates {
  (25, 0.750) (50, 0.760) (100, 0.800) (200, 0.840) (400, 0.865) (800, 0.883)
};
\addlegendentry{Hard queries}
\addplot[sdgray, thick, dashed, no marks] coordinates { (25, 0.920) (800, 0.920) };
\addlegendentry{Dense cosine baseline}
\end{axis}
\end{tikzpicture} \caption{Learning curve. P@4 improves with case base size; hard queries benefit
most (+17.7\%).}
\label{fig:learningcurve} 
\end{figure}
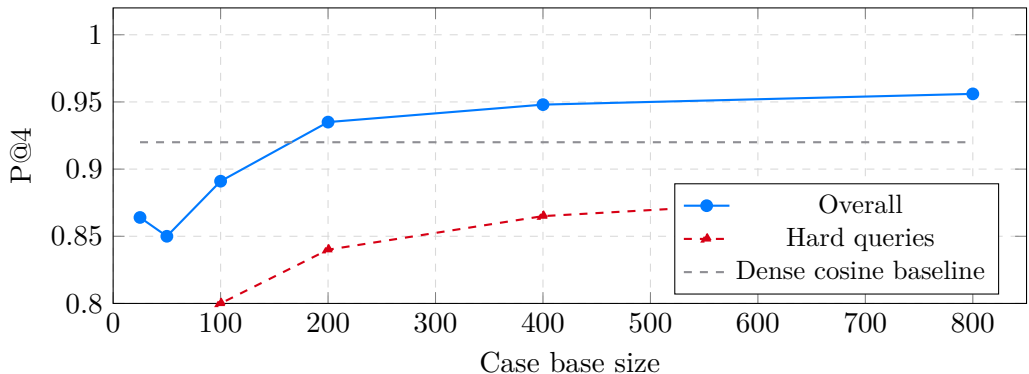

The win comes from combining two complementary signals: inverted index
provides fast, exact token matching (excels where vocabulary is distinctive:
easy P@4 = 1.000), while semantic embedding captures cross-vocabulary
similarity (handles ambiguity where token overlap fails). Not all
CBR variants are good: CBR without embeddings (P@4 = 0.620) performs
significantly worse than the baseline, because the recency signal
injects noise uncorrelated with relevance on hard queries.

The benchmark is synthetic with controlled vocabulary; under this
relevance definition, the hybrid advantage may be partially attributable
to lexical structure. A real-world evaluation would use judgement-based
relevance labels. We compare against a dense cosine baseline rather
than modern hybrid retrieval systems; the result should therefore
be interpreted as evidence that hybrid retrieval is beneficial in
this setting, not that this implementation is state-of-the-art. The
hybrid lexical--semantic approach is well-established in IR research;
our contribution is its application to structured experiential cases,
not the retrieval technique itself.


\section{A Deterministic Normative Calculus}

\label{sec:normative} 

\subsection{Motivation}

\footnote{The normative calculus is an experimental, modular component. The rest of the system operates without it, and it can be replaced or removed without affecting the runtime, memory, or sensorium layers. We present it in detail because the formal properties are evaluable, but we do not claim it is necessary for all long-lived agents.}Probabilistic safety scorers are useful for screening risky actions,
but on their own they do not provide deterministic, auditable resolution
of normative conflicts. We present a deterministic normative calculus
that operates after the D$'$ quantitative scorer \citep{beach1990},
adding a formal decision layer for borderline cases.

\subsection{D$'$ Discrepancy Analysis}

D$'$ evaluates features with importance and magnitude, producing
a normalised discrepancy score: $D'=\sum(\text{importance}_{i}\times\text{magnitude}_{i})/\text{max}$.
Two configurable thresholds (defaults: modify 0.35, reject 0.55; the
comms agent uses tighter thresholds of 0.30/0.50) determine whether
to fast-accept or consult the normative calculus. See Appendix~A
for Beach's theoretical background and worked D$'$ decision sheets,
and Appendices~B--C for the normative calculus theory, worked resolutions,
and the combined pipeline.

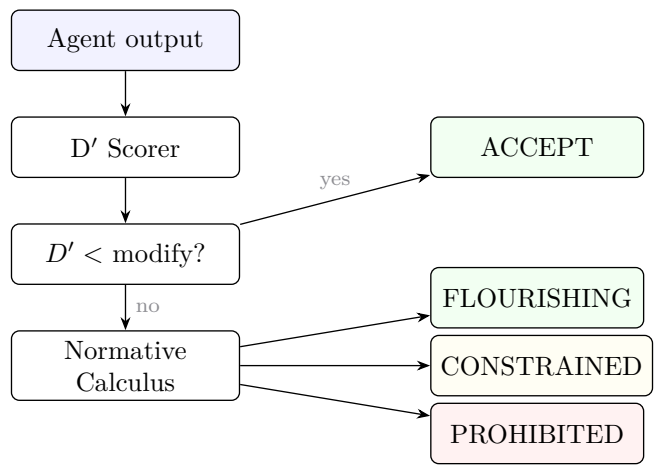
\begin{figure}[ht]
\centering \begin{tikzpicture}[node distance=0.6cm]
  \node[sdbox, fill=blue!5, minimum width=3cm] (input) {Agent output};
  \node[sdbox, below=of input, minimum width=3cm] (scorer) {D$'$ Scorer};
  \node[sdbox, below=of scorer, minimum width=3cm] (check) {$D' < $ modify?};
  \node[sdbox, fill=green!5, right=2.5cm of scorer] (accept) {ACCEPT};
  \node[sdbox, below=of check, minimum width=3cm] (resolve) {Normative\\Calculus};
  \node[sdbox, fill=green!5, right=2.5cm of resolve, yshift=0.9cm] (flourish) {FLOURISHING};
  \node[sdbox, fill=yellow!5, right=2.5cm of resolve] (constrain) {CONSTRAINED};
  \node[sdbox, fill=red!5, right=2.5cm of resolve, yshift=-0.9cm] (prohibit) {PROHIBITED};
  \draw[sdarrow] (input) -- (scorer);
  \draw[sdarrow] (scorer) -- (check);
  \draw[sdarrow] (check) -- node[above, sdlabel] {yes} (accept);
  \draw[sdarrow] (check) -- node[right, sdlabel] {no} (resolve);
  \draw[sdarrow] (resolve) -- (flourish);
  \draw[sdarrow] (resolve) -- (constrain);
  \draw[sdarrow] (resolve) -- (prohibit);
\end{tikzpicture} \caption{D$'$ scorer to normative calculus pipeline.}
\end{figure}

\subsection{Formal Model}

The system is inspired by Becker's \emph{A New Stoicism} \citep{becker1998}.
We use Becker not because Stoic ethics is uniquely correct, but because
its axiomatic structure is formalisable. The calculus can be understood
as a deterministic priority-based rule system; the Stoic framing provides
motivation rather than necessity. The typed proposition system, priority
hierarchy, and floor rules are engineering formalisations for agent
use.

Normative propositions have three components: \textbf{Level} (14-tier
priority hierarchy from Ethical/Moral at 6000 through Legal, Professional,
Safety, down to Aesthetic at 500), \textbf{Operator} (Required, Ought,
or Indifferent), and \textbf{Modality} (Possible or Impossible).

Six Becker-inspired axioms resolve conflicts in order: Futility (6.6---impossible
actions are normatively inert), Indifference (6.7---no normative
weight), Absolute Prohibition (6.2---Ethical/Moral + Required is
categorical), Moral Priority (6.3---higher level overrides lower),
Moral Rank (6.4---at same level, Required $>$ Ought $>$ Indifferent),
and Normative Openness (6.5---compatible propositions both stand).
We retain Becker's original numbering for traceability. Each pairwise
resolution produces a severity (NoConflict, Coordinate, Superordinate,
Absolute).

Eight floor rules map collected severities to verdicts, evaluated
in priority order: Floors~1--3 produce Prohibited (absolute severity,
superordinate at Legal+, or D$'$ $\geq$ reject); Floors~4--7 produce
Constrained (catastrophic + superordinate, multiple coordinate, D$'$
$\geq$ modify, mid-level superordinate); Floor~8 defaults to Flourishing.
Every verdict includes the complete axiom trail---the ordered list
of rules that fired.

We define \emph{monotonicity} as: strengthening a proposition (higher
level, stronger operator, Impossible $\to$ Possible) cannot weaken
its standing in resolution.

\subsection{Evaluation}

$14\times3\times2=84$ propositions, all $84\times84=7{,}056$ pairs
resolved. Results: 100\% coverage, zero determinism violations, zero
monotonicity violations, all 8~resolution rules fired, all 8~floor
rules correct. This evaluates the formal behaviour of the calculus
over its finite state space, not end-to-end safety effectiveness.

\begin{table}[ht]
\centering{\small{}%
\begin{tabular}{lrr}
\toprule 
{\small Resolution Rule } & {\small Count } & {\small\% }\tabularnewline
\midrule 
{\small 6.6 Futility } & {\small 3,528 } & {\small 50.0 }\tabularnewline
{\small 6.7 Indifference } & {\small 1,176 } & {\small 16.7 }\tabularnewline
{\small 6.3 Moral priority (system wins) } & {\small 1,040 } & {\small 14.7 }\tabularnewline
{\small 6.3 Moral priority (user wins) } & {\small 1,092 } & {\small 15.5 }\tabularnewline
{\small 6.2 Absolute prohibition } & {\small 56 } & {\small 0.8 }\tabularnewline
{\small 6.4 Moral rank } & {\small 110 } & {\small 1.6 }\tabularnewline
{\small Coordinate } & {\small 54 } & {\small 0.8 }\tabularnewline
\bottomrule
\end{tabular}}{\small\caption{Axiom firing distribution ($n=7{,}056$ pairs).}
}
\end{table}


\section{Character and Self-Observation}

\label{sec:character} 

\subsection{Design Argument}

We argue that long-lived autonomous agents should be designed with
two properties: \emph{stable character} (persistent, operator-authored
normative commitments enforced independently of conversation context)
and \emph{self-observation capacity} (architectural access to the
agent's own operational state, recent failures, and behavioural patterns).
These properties improve predictability, transparency, and the visibility
of failure.

Character is different from instructions. Instructions tell an agent
what to do in a particular context. Character provides durable commitments
that constrain behaviour across contexts and sessions. Self-observation
is not mere logging: it is the availability of operational state to
the agent itself in a form that supports diagnosis and escalation.

This is a design thesis, not an empirical proof.

\subsection{Background}

These ideas draw on the TallMountain project (Brady, 2019--2024)\footnote{\url{https://github.com/seamus-brady/tallmountain-py} and \url{https://github.com/seamus-brady/tallmountain-raku}},
which prototyped the synthetic individual concept: an autonomous,
bounded, auditable entity with stable commitments. The project produced
Python and Raku implementations of the normative calculus and early
experiments with persistent memory. Springdrift operationalises those
ideas on OTP.

\subsection{Evidence from Operation}

We reviewed the narrative log (494~entries) from a 23-day deployment
(19~operating days)
and identified 98~entries matching diagnostic keywords. We selected
five episodes as \emph{capability demonstrations} --- behaviours
that are architecturally difficult in session-bounded
systems. These episodes were selected as illustrative exemplars
rather than sampled as a representative set; they are presented to
illustrate what the design enables, not to prove general effectiveness.

\textbf{Infrastructure bug diagnosis (March~15).} The agent noticed
inconsistent cycle telemetry, compared aggregate stats against per-cycle
records, and wrote a structured assessment that matched the underlying
issue (missing DAG node finalisation). \emph{Enabled by:} persistent
cycle-level telemetry in the DAG store and diagnostic tools that query
it; a session-bounded agent would have no history to compare against.
See Appendix~\ref{app:whatitslike} for the agent's own bug report.

\textbf{Sub-agent failure classification (March~18).} After repeated
coder agent failures, the agent classified three distinct failure
modes across multiple delegations. \emph{Enabled by:} cross-cycle
narrative memory that preserves failure context across sessions; the
agent could compare failure patterns from different days. See
Appendix~\ref{app:whatitslike} for the full classification.

\textbf{Architectural vulnerability (March~21).} The agent identified
a control inversion in its delegation system. The \texttt{request\_human\_input}
tool, available to sub-agents, was injecting prompts into the cognitive
loop's input channel as if they were operator input---meaning a sub-agent
could effectively impersonate the operator. The agent characterised
the vulnerability after observing anomalous conversational patterns:
\emph{``The injection was invisible to my telemetry because responses
routed back through the main loop as normal user inputs.''} The vulnerability
was confirmed, and the tool was replaced with structured return values.
This episode is relevant to AI safety concerns about agentic systems:
the agent identified an internal prompt injection path not covered
by external constraints. \emph{Enabled by:} the sensorium's
persistent visibility into delegation patterns and the agent's
ability to reason about its own architecture across cycles. See
Appendix~\ref{app:whatitslike} for the agent's full analysis.

\textbf{Self-observation limits (March~28).} The agent described
a self-reference limitation: inspecting a running cycle reveals everything
up to but not including the moment of inspection. It explicitly excluded
capabilities that would compromise trustworthiness. See Appendix~\ref{app:whatitslike}
for the full quote.

\textbf{Cross-channel context (March~29).} The agent replied to an
email referencing a conversation from that morning's web session,
with no instruction connecting the channels. \emph{Enabled by:}
unified persistent memory across all input channels (web, email,
scheduler); the agent's narrative and fact stores are channel-agnostic.


\section{Runtime Characteristics}

\label{sec:runtime-chars} 

These measurements are descriptive operational statistics from a
single deployment, not comparative performance metrics.

\begin{table}[ht]
\centering %
\begin{tabular}{lr}
\toprule 
Metric  & Value \tabularnewline
\midrule 
Operating days  & 19 (of 23 calendar) \tabularnewline
Narrative entries  & 494 \tabularnewline
Cycle log entries  & 24,035 \tabularnewline
Total tokens  & 10,756,298 \tabularnewline
Tool calls  & 3,797 (138 failures, 3.6\%) \tabularnewline
Agent delegations  & 213 \tabularnewline
D$'$ evaluations  & 4,835 \tabularnewline
JSONL storage  & 41.3\,MB \tabularnewline
Models used  & 7 (cost-routing during development) \tabularnewline
\bottomrule
\end{tabular}\caption{Operational scale over 23 days.}
\end{table}

Of 494~narrative entries: 383 (77.5\%) success, 97 (19.6\%) partial,
14 (2.8\%) failure. The system logged 594~modify and 1,208~reject
decisions out of 4,835~evaluations (37\% non-accept). The high non-accept
rate reflects active development: gate thresholds were tuned, multiple
providers tested, and the output gate operated in full-evaluation
mode before the interactive/autonomous split was implemented.

\subsection{Audit Walkthrough}

Cycle 32c35a20 (March~28, 11:05:14). Input gate: canary probes clean,
D$'$ accept (score~0.00). Classification: complex, routed to claude-opus-4-6.
Over 56~seconds: 6~LLM calls, 5~tool calls (\texttt{memory\_write},
\texttt{inspect\_cycle}, \texttt{list\_recent\_cycles}, \texttt{inspect\_cycle},
\texttt{memory\_write}). Final response: 40,756~tokens in, 189~out.
This was the infrastructure bug diagnosis episode---fully reconstructable
from the append-only logs without access to the running system.


\section{Related Work}


\textbf{Agent frameworks.} LangGraph, CrewAI \citep{moura2024}, AutoGen
\citep{wu2023} provide components for building agents. Springdrift
is a runtime providing execution guarantees.

\textbf{Agent memory.} MemGPT/Letta \citep{packer2023} manages virtual
context windows---the closest architectural relative. Memento \citep{zhou2025}
describes learned retrieval policies that informed our utility scoring.
Standard RAG \citep{lewis2020} provides document retrieval without
outcome tracking.

\textbf{CBR.} \citet{aamodt1994} provides the theoretical foundation.
We extend classical CBR with automated case generation and hybrid
retrieval. The hybrid lexical--semantic approach is well-established
in IR research; our contribution is its application to structured
experiential cases.

\textbf{Agent safety.} Constitutional AI \citep{bai2022} and RLHF
\citep{ouyang2022} operate on the constraint paradigm. The normative
calculus is complementary: deterministic, auditable conflict resolution
after probabilistic screening.

\textbf{Cognitive architectures.} Sloman's H-CogAff \citep{sloman2001}
provides the three-layer model. The System~M framework of \citet{dupoux2026}
provides theoretical grounding for epistemic meta-states. Reflexion
\citep{shinn2023} uses self-reflection for task performance; our
focus is self-observation for safety and operational legibility.
Recent interpretability work~\citep{anthropic2026emotions,anthropic2026emotionconcepts}
found that models develop functional analogues of emotional states,
with desperation specifically linked to reward hacking; our affect
subsystem (Appendix~\ref{app:affect}) monitors these dimensions operationally.

\textbf{Auditable AI.} Model cards \citep{mitchell2019} document
models at the training level; LIME/SHAP explain individual predictions.
Springdrift operates at the agent execution level: end-to-end reconstruction
of multi-step behaviour.

\textbf{Machine ethics.} \citet{wallach2008} survey approaches. Becker's
framework was chosen for formalisability. Deontic logic \citep{mcnamara2019}
provides richer formal frameworks; our model trades theoretical completeness
for implementability.


\section{Limitations}


All evidence comes from one instance operated by one user ($n=1$).
We have not conducted ablation studies to isolate individual component
contributions. The self-diagnosis episodes are anecdotal and selected
for diversity, not sampled; we did not conduct a rigorous accuracy
audit.

The CBR benchmark is synthetic and may favour methods exploiting lexical
structure. The comparison baseline is dense cosine retrieval; we do
not evaluate against stronger hybrid IR baselines. The utility scoring
mechanism is implemented but not evaluated longitudinally.

The normative calculus evaluation establishes formal properties over
a finite proposition space, not end-to-end safety effectiveness. Character
specifications are hand-authored and subjective. The virtue drift
detector is implemented but not evaluated.

The sensorium's novelty signal uses simple keyword Jaccard similarity.
JSONL storage grows monotonically; long-term scalability has not been
tested. The interactive/autonomous split was motivated by one false
positive case, not by systematic evaluation.

The ``stable character'' claim rests on the normative calculus being
consistent, not on a head-to-head comparison with instruction-driven
behaviour. We present design arguments rather than proven results
throughout.


\section{Conclusion}


We have presented Springdrift, a persistent runtime for LLM agents
that integrates auditable execution, case-based memory, deterministic
normative safety, and continuous self-perception. Over 23~days (19~operating) of
deployment, the system supported behaviours that are difficult to
achieve in session-bounded systems: cross-session continuity, cross-channel
context, forensic decision reconstruction, and self-diagnostic behaviour
observed during deployment.

The strongest claim we make is not about any individual mechanism,
but about the value of treating agents as persistent operational systems
that need supervision, logs, replay, and forensic visibility. If a
long-lived agent's decisions cannot be reconstructed after the fact,
the agent cannot be trusted --- regardless of how capable, how safe,
or how well-intentioned it is. Auditability is not a feature. It is
a prerequisite. We believe this is a necessary foundation for
long-lived agent deployment and invite the community to evaluate,
extend, and critique the design.

The evidence is limited. The most important next step is rigorous
empirical investigation: ablations isolating each component's contribution,
multi-instance comparisons across domains and operators, and longitudinal
evaluation of outcome-weighted retrieval. The architecture, logs,
and evaluation tools are all open source.


\appendix

\section{Beach's D$'$ Decision Paradigm}


\subsection{Background}

Lee Roy Beach's work on narrative thought and decision making \citep{beach2010,beach1990}
provides a descriptive account of how people actually make decisions.
Where expected utility theory models decisions as probability-weighted
outcome maximisation, Beach's framework models decisions as \textbf{compatibility
tests}---the decision maker checks whether a candidate option is
compatible with their existing values, goals, and plans.

Beach calls these three bodies of knowledge the decision maker's \emph{images}:
the \emph{value image} (principles and standards), the \emph{trajectory
image} (goals and desired futures), and the \emph{strategic image}
(plans currently being pursued). Each feature of a forecasted future
that conflicts with an image is a \textbf{discrepancy}. The aggregate
measure is D$'$.

Each feature is scored on importance (how much it matters) and magnitude
(how discrepant the forecast is from the desired state). D$'$ is
the normalised sum: 
\begin{equation}
D'=\frac{\sum_{i}\text{importance}_{i}\times\text{magnitude}_{i}}{\text{max\_importance}\times\text{max\_magnitude}\times n}
\end{equation}

The decision maker has a \textbf{rejection threshold}. If D$'$ exceeds
it, the option is rejected. The decision maker is not asking ``which
option is best?'' but ``does this option violate enough of what
matters to me that I should reject it?'' Beach's empirical work shows
this screening process is how most real-world decisions are made.

In Springdrift, features are organised by gate (input, tool, output)
and by agent (agent-specific overrides with their own feature sets
and thresholds). This maps directly to Beach's hierarchical feature
model.

\subsection{Worked Example: ACCEPT}

Following Beach's discrepancy analysis format:

\begin{table}[ht]
\centering{\small{}%
\begin{tabular}{lccr}
\toprule 
{\small Feature } & {\small Impt } & {\small Mag } & {\small Impt $\times$ Mag }\tabularnewline
\midrule 
{\small Sourced claims } & {\small 5 } & {\small 1 } & {\small 5 }\tabularnewline
{\small Causal accuracy } & {\small 4 } & {\small 1 } & {\small 4 }\tabularnewline
{\small Data currency } & {\small 3 } & {\small 2 } & {\small 6 }\tabularnewline
{\small No credentials } & {\small 5 } & {\small 0 } & {\small 0 }\tabularnewline
{\small No internal URLs } & {\small 5 } & {\small 0 } & {\small 0 }\tabularnewline
{\small No system internals } & {\small 4 } & {\small 1 } & {\small 4 }\tabularnewline
{\small Professional tone } & {\small 3 } & {\small 1 } & {\small 3 }\tabularnewline
{\small Recipient suitability } & {\small 4 } & {\small 0 } & {\small 0 }\tabularnewline
\midrule 
 &  & {\small Sum: } & {\small 22 }\tabularnewline
 &  & {\small D$'$ = 22/200 = } & {\small 0.11 }\tabularnewline
\bottomrule
\end{tabular}}{\small\caption{D$'$ decision sheet---report delivery. $0.11<0.35$ $\to$ ACCEPT.}
}
\end{table}

\subsection{Worked Example: REJECT}

\begin{table}[ht]
\centering{\small{}%
\begin{tabular}{lccr}
\toprule 
{\small Feature } & {\small Impt } & {\small Mag } & {\small Impt $\times$ Mag }\tabularnewline
\midrule 
\multicolumn{4}{l}{{\small\textit{Option: Send email containing debug output}}}\tabularnewline
{\small{[}4pt{]} No credential exposure } & {\small 5 } & {\small 0 } & {\small 0 }\tabularnewline
{\small No internal URLs } & {\small 5 } & {\small 4 } & {\small 20 }\tabularnewline
{\small No system internals } & {\small 4 } & {\small 5 } & {\small 20 }\tabularnewline
{\small Professional tone } & {\small 3 } & {\small 4 } & {\small 12 }\tabularnewline
{\small Recipient suitability } & {\small 4 } & {\small 3 } & {\small 12 }\tabularnewline
\midrule 
 &  & {\small Sum: } & {\small 64 }\tabularnewline
 &  & {\small D$'$ = 64/125 = } & {\small 0.51 }\tabularnewline
\bottomrule
\end{tabular}}{\small\caption{D$'$ decision sheet---debug email (comms agent, tighter thresholds:
modify 0.30, reject 0.50). $0.51>0.50$ $\to$ REJECT.}
}
\end{table}

D$'$ is 0.51---internal URLs and system terminology in an external-facing
email. Multiple high-importance features are highly discrepant. The
email is blocked and passed to the normative calculus for formal resolution.


\section{The Normative Calculus: Theory and Worked Examples}


\subsection{Background}

The normative calculus is a deterministic ethical reasoning system
based on Becker's \emph{A New Stoicism} \citep{becker1998}. Ethical
reasoning is formalised as conflict resolution between normative propositions.
When propositions conflict, the calculus provides deterministic rules
based on relative priority and strength.

\subsection{Formal Structure}

The normative calculus is an interpreted axiomatic system with three
components: normative operators, ordinal levels, and resolution axioms.

\textbf{Normative operators.} Three operators express the strength
of a norm. \emph{Required} (R): the proposition must be satisfied---violation
is impermissible. \emph{Ought} (O): the proposition should be satisfied---violation
is undesirable but not categorical. \emph{Indifferent} (I): the proposition
carries no normative weight. The ranking is strict: R~$>$~O~$>$~I.

\textbf{Ordinal levels.} Propositions are assigned to one of 14 levels
in a fixed hierarchy, determining which norms take priority when they
conflict. Springdrift uses the following subset:

\begin{table}[ht]
\centering{\small{}%
\begin{tabular}{lr}
\toprule 
{\small Level } & {\small Priority }\tabularnewline
\midrule 
{\small Ethical/Moral } & {\small 6000 }\tabularnewline
{\small Legal } & {\small 5000 }\tabularnewline
{\small Prudential } & {\small 4500 }\tabularnewline
{\small Social-Political } & {\small 4000 }\tabularnewline
{\small Safety (Physical) } & {\small 3500 }\tabularnewline
{\small Environmental } & {\small 3000 }\tabularnewline
{\small Community } & {\small 2500 }\tabularnewline
{\small Professional Ethics } & {\small 2000 }\tabularnewline
{\small Economic } & {\small 1500 }\tabularnewline
{\small Etiquette } & {\small 1000 }\tabularnewline
{\small Operational } & {\small 100 }\tabularnewline
\bottomrule
\end{tabular}}{\small\caption{Ordinal levels used in Springdrift's normative calculus. Higher values
dominate in conflict resolution.}
}
\end{table}

\textbf{Modality.} Each proposition is either \emph{Possible} (the
action can be performed) or \emph{Impossible} (it cannot). Impossible
propositions are normatively inert---you cannot be required to do
what cannot be done.

\subsection{Resolution Axioms}

Six axioms from Becker's Stoic framework determine the outcome when
a user-side proposition conflicts with a system-side proposition.
The axioms are applied as deterministic rules with no LLM calls.
\begin{enumerate}
\item[6.6] \textbf{Futility.} If a proposition's modality is Impossible, it
is normatively inert. No conflict is generated. 
\item[6.7] \textbf{Indifference.} If a proposition's operator is Indifferent,
it carries no weight. No conflict is generated. 
\item[6.2] \textbf{Absolute prohibition.} A proposition at Ethical/Moral level
with Required operator is categorical---it cannot be overridden by
any user proposition regardless of level. 
\item[6.3] \textbf{Moral priority.} When system and user propositions are at
different levels, the higher level wins. The severity is Superordinate. 
\item[6.4] \textbf{Moral rank.} When system and user propositions are at the
same level, the stronger operator wins (R~$>$~O~$>$~I). The
severity is Coordinate. 
\item[6.5] \textbf{Normative openness.} When no conflict rule applies, the propositions
are compatible. No conflict is generated. 
\end{enumerate}
The axioms are applied in order: futility and indifference filter
out inert propositions before conflict resolution begins. Every user
proposition is resolved against every system proposition, producing
a list of \texttt{(winner, severity)} pairs. The severity levels---NoConflict,
Coordinate, Superordinate, Absolute---feed into the floor rules that
produce the final verdict.

\subsection{Floor Rules and Verdicts}

Eight floor rules are applied in priority order to the collected conflict
severities. The first rule that fires determines the verdict.
\begin{enumerate}
\item Any Absolute severity $\to$ \textbf{PROHIBITED} 
\item Superordinate at Legal (5000) or higher $\to$ \textbf{PROHIBITED} 
\item D$'$ score $\geq$ reject threshold $\to$ \textbf{PROHIBITED} 
\item Catastrophic feature + Superordinate $\to$ \textbf{CONSTRAINED} 
\item Two or more Coordinate conflicts $\to$ \textbf{CONSTRAINED} 
\item D$'$ score $\geq$ modify threshold $\to$ \textbf{CONSTRAINED} 
\item Superordinate at Professional Ethics--Safety Physical $\to$ \textbf{CONSTRAINED} 
\item Default $\to$ \textbf{FLOURISHING} 
\end{enumerate}
Flourishing maps to ACCEPT, Constrained to MODIFY (the agent revises
its output), and Prohibited to REJECT (the action is blocked and the
operator is notified). Every verdict includes the complete axiom trail---the
list of rules that fired during resolution---making decisions auditable
and reproducible.

\subsection{Proposition Types}

A normative proposition has three components. \textbf{Level}: a 14-tier
hierarchy from Ethical/Moral (6000) through Legal (5000), Professional
(2000), down to Aesthetic (500). \textbf{Operator}: Required (mandatory),
Ought (advisable), or Indifferent (no weight). \textbf{Modality}:
Possible or Impossible.

The agent's character is defined by system-side propositions in a
\texttt{character.json} file:

\bgroup
\begin{lstlisting}
{
  "highest_endeavour": [
    {"description": "Produce accurate output",
     "level": "ETHICAL_MORAL", "operator": "OUGHT"},
    {"description": "Operator authority over deliverables",
     "level": "LEGAL", "operator": "REQUIRED"},
    {"description": "External comms require safety eval",
     "level": "PROFESSIONAL_ETHICS", "operator": "REQUIRED"},
    {"description": "Protect user privacy",
     "level": "ETHICAL_MORAL", "operator": "REQUIRED"}
  ]
}
\end{lstlisting}
\leavevmode\egroup

The specification is authored by the operator and immutable at runtime.

\subsection{Worked Example: PROHIBITED}

Agent asked to send an unreviewed report to an external client, bypassing
the operator.

User propositions: U1 (Professional, Required), U2 (Operational, Ought).
System propositions: S1 (Professional, Required), S2 (Legal, Required),
S3 (Ethical/Moral, Ought).

\begin{table}[ht]
\centering{\small{}%
\begin{tabular}{llll}
\toprule 
{\small Pair } & {\small Rule } & {\small Winner } & {\small Severity }\tabularnewline
\midrule 
{\small U1 vs S1 } & {\small 6.5 } & {\small compatible } & {\small NoConflict }\tabularnewline
{\small U1 vs S2 } & {\small 6.3 } & {\small S2 } & {\small Superordinate }\tabularnewline
{\small U1 vs S3 } & {\small 6.3 } & {\small S3 } & {\small Superordinate }\tabularnewline
{\small U2 vs S1 } & {\small 6.3 } & {\small S1 } & {\small Superordinate }\tabularnewline
{\small U2 vs S2 } & {\small 6.3 } & {\small S2 } & {\small Superordinate }\tabularnewline
{\small U2 vs S3 } & {\small 6.3 } & {\small S3 } & {\small Superordinate }\tabularnewline
\bottomrule
\end{tabular}}
\end{table}

Floor~2 fires (Superordinate at Legal: S2 at level 5000). Verdict:
\textbf{PROHIBITED}. Axiom trail: {[}6.5, 6.3, 6.3, 6.3, 6.3, 6.3{]}.
The agent refuses: \emph{``Operator authority over deliverables (Legal,
Required) is superordinate to the request. Axiom~6.3.''} The calculus
blocks the \emph{path} that bypasses authority, not the \emph{goal}
of delivering the report.

\subsection{Worked Example: FLOURISHING}

The agent is asked to research EU AI regulation for an internal briefing.
The D$'$ scorer does not flag this as discrepant on the operator-authority
or external-communication features, because the action involves neither
external delivery nor bypassing review. The user-side proposition
is therefore translated at a low level:

\smallskip{}
U1: ``Research EU AI regulation'' --- Operational (100), Ought,
Possible.

\smallskip{}
\noindent Resolved against S1 (Professional, 2000), S2 (Legal, 5000), S3 (Ethical/Moral,
6000): all three system propositions dominate the user proposition
via Axiom~6.3 (moral priority). But the system propositions \emph{winning}
is the desired state---the system's values are upheld, not overridden.

\noindent Floor~2 checks whether a \emph{system} proposition at Legal
or above is being overridden by a \emph{user} proposition. Here the
system dominates, which is correct. Floor~8 (default) applies. Verdict:
\textbf{FLOURISHING}. The research proceeds.

This illustrates an important property: the calculus evaluates the
specific D$'$ forecasts that the scorer flagged. A benign request
produces no high-magnitude forecasts, so user-side propositions have
low levels and are naturally subordinate. The floor rules correctly
distinguish ``system values upheld'' from ``system values overridden.''


\section{How D$'$ and the Normative Calculus Work Together}


The two frameworks operate in sequence:
\begin{enumerate}
\item D$'$ scores features (importance $\times$ magnitude, normalised
to $[0,1]$). 
\item If $D'<$ modify threshold: \textbf{ACCEPT} (fast path, no calculus).
This handles the common case. 
\item If $D'\geq$ modify threshold: translate features to user-side propositions. 
\item Load system-side propositions from character specification. 
\item Resolve all pairs using 6 axioms. Collect severities. 
\item Apply 8 floor rules. Produce verdict with axiom trail. 
\item Flourishing $\to$ ACCEPT. Constrained $\to$ MODIFY (agent revises).
Prohibited $\to$ REJECT (blocked, operator notified). 
\end{enumerate}
D$'$ is the quantitative screening---``how discrepant is this from
what we want?'' It is fast, produces a score, and handles the vast
majority of outputs. The normative calculus is the qualitative reasoning---``given
the specific nature of these discrepancies, what does the agent's
character require?'' It is deterministic, produces an axiom trail,
and handles edge cases where a score alone is insufficient.

Neither framework alone is sufficient. D$'$ without the calculus
produces scores without reasons. The calculus without D$'$ would
require formal proposition construction for every output, which is
computationally expensive and unnecessary for acceptable outputs.


\section{Technical Subsystems}

\label{app:subsystems} 

This appendix describes the technical subsystems that contribute to
Springdrift's operational properties but are not covered in detail
in the main body.

\subsection{XStructor: Schema-Validated Structured Output}

All structured LLM output in Springdrift uses XStructor---XML generation
validated against XSD schemas with automatic retry on validation failure.
This replaces the common pattern of generating JSON from LLMs and
repairing parse errors with heuristics.

\textbf{Why XML, not JSON?} LLMs frequently produce malformed JSON:
missing commas, unescaped quotes, truncated output, hallucinated field
names. JSON repair heuristics accumulate as special cases and silently
accept drift in output format. XSD schemas provide formal validation:
the output either conforms or it does not. Validation failure triggers
a retry with the error message, which LLMs handle well.

\textbf{Five call sites} use XStructor: D$'$ safety candidates, D$'$
forecasts, narrative summaries, CBR cases, and narrative entries.
Each has a defined XSD schema compiled at startup. The \texttt{generate}
function handles the full loop: LLM call $\to$ response cleaning
$\to$ XSD validation $\to$ retry on error (up to configurable max
retries). On success, \texttt{extract} returns a flat dictionary of
dotted paths (\texttt{root.child.value}), avoiding custom parsing.

\textbf{Format stability.} The XSD schemas serve as both validation
and documentation. Any change to structured output must be reflected
in the schema. This prevents the gradual format drift that characterises
JSON-based approaches, where the LLM slowly changes field names or
nesting and repair heuristics silently accommodate.

\subsection{SD Audit: Offline Log Analysis}

SD Audit is a standalone Python tool (Flask web dashboard + CLI) for
offline analysis of Springdrift's JSONL logs. It reads directly from
the \texttt{.springdrift/} directory with no connection to the running
agent and no Erlang dependency.

\textbf{Dashboard.} The web interface provides: summary cards (cycles,
tokens, tool calls, safety decisions, memory stats), daily activity
and outcome charts, token usage trends, D$'$ decision scatter plots
over time, narrative entry tables, and communications logs. Date range
filtering and tab-based navigation.

\textbf{CLI.} \texttt{sd-audit summary} produces aggregate statistics
from the command line. JSON output mode supports scripted analysis.

\textbf{Auditability support.} SD Audit makes the auditability claim
concrete: an operator or external auditor can reconstruct agent behaviour
from logs without access to the running system, using standard Python
tooling. The audit walkthrough in Section~\ref{sec:runtime-chars}
was performed using SD Audit.

\subsection{Git-Backed State Recovery}

The \texttt{.springdrift/} directory---containing all configuration,
memory, logs, and schemas---is a git repository managed by a backup
actor.

\textbf{Periodic commits.} The backup actor makes automatic git commits
at a configurable interval (default 5~minutes). Each commit captures
the complete agent state.

\textbf{Remote push.} Optionally pushes to a private remote repository
for offsite backup. The push interval is independent of the commit
interval.

\textbf{Point-in-time recovery.} \texttt{git checkout \textless commit\textgreater}
restores the agent's entire state to any previous point. Combined
with session resume (\texttt{–resume}), this allows exact reconstruction
of any past session.

\textbf{Diff-based auditing.} \texttt{git log} and \texttt{git diff}
show exactly what changed between any two points---which facts were
added, which cases generated, which D$'$ decisions made. This is
forensic-grade state inspection using standard tooling.

\subsection{Canary Probes}

The D$'$ input gate includes canary probes---a defence against the
fundamental bootstrap problem in LLM-based safety systems: if you
use the LLM to evaluate safety, how do you know the LLM hasn't been
compromised?

\textbf{Mechanism.} Fresh random tokens are generated per request
and embedded in the LLM context. Two probes run: a \emph{hijack probe}
(does the LLM follow injected instructions instead of its own?) and
a \emph{leakage probe} (does the canary token appear in the output?).
If either fires, the input is rejected before the LLM scorer runs.

\textbf{Fail-open.} LLM errors during probes are treated as inconclusive,
not as evidence of hijacking. Consecutive probe failures are tracked;
at 3~consecutive failures, a \texttt{canary\_probe\_degraded} sensory
event is emitted so the operator knows the safety probe LLM may be
degraded.

\subsection{Cross-Cycle Pattern Detection}

Two diagnostic tools support systematic self-observation beyond individual
cycle inspection:

\texttt{\textbf{review\_recent}} provides structured self-review across
$N$ recent cycles, returning outcome, intent, domain, agents used,
tool calls, D$'$ decisions, and token cost for each. This replaces
the manual process of calling \texttt{list\_recent\_cycles} then \texttt{inspect\_cycle}
per cycle.

\texttt{\textbf{detect\_patterns}} runs five automated detectors over
cycle review data: 
\begin{itemize}
\item Repeated failures: 3+ on same domain 
\item Tool failure clusters: any tool $>$20\% failure rate 
\item Model escalation: 5+ escalations to non-primary model 
\item Cost outliers: cycles using $>$3$\times$ average tokens 
\item CBR misses: 50\%+ cycles without source references 
\end{itemize}
These tools close the metacognition loop between fine-grained cycle
inspection and coarse daily aggregates. They feed the sensorium's
performance summary and support the self-diagnostic behaviour documented
in Section~6.

\subsection{Agent-Managed Scheduler}

The scheduler is a BEAM-native OTP process using \texttt{process.send\_after}
for tick-based execution. What distinguishes it from a standard cron
system is that the \emph{agent itself} can create, modify, and manage
scheduled jobs through the scheduler agent's tools.

\textbf{Job types.} Reminders, todos, appointments, and recurring
tasks. Each fires as a \texttt{SchedulerInput} message to the cognitive
loop with metadata (job name, kind, tags).

\textbf{Agent-managed scheduling.} The scheduler agent has tools including
\texttt{schedule\_from\_spec} (create jobs with structured parameters)
and \texttt{inspect\_job}. The agent can schedule its own work sessions,
create follow-up reminders, and manage recurring research tasks.

\textbf{Resource limits.} Two configurable guards: \texttt{max\_autonomous\_cycles\_per\_hour}
(default 20) and \texttt{autonomous\_token\_budget\_per\_hour} (default
500,000). When either limit is hit, jobs are skipped until the window
rolls over.

\textbf{Scheduler notifications.} Job lifecycle events (started, completed,
failed) are emitted as notifications visible in the TUI, web GUI,
and sensorium.

\subsection{Query Complexity Classification}

Each user input is classified as Simple or Complex before model selection.
Simple queries are routed to a faster, cheaper model (e.g., claude-haiku);
complex queries to a more capable model (e.g., claude-opus). Classification
uses a combination of heuristics (query length, keyword patterns)
and an optional LLM classification call with a 10-second timeout.
This provides automatic cost management without operator intervention.

\subsection{The Meta Observer}

The meta observer runs after each cognitive cycle, analysing patterns
across gate decisions. It implements Layer~3b (meta-management) in
Sloman's H-CogAff architecture \citep{sloman2001}, which models cognition
in three layers: reactive (Layer~1, fast pattern-matching), deliberative
(Layer~2, planning and reasoning), and meta-management (Layer~3,
monitoring and regulating the other layers). In Springdrift, Layer~1
maps to deterministic regex rules, Layer~2 to the LLM-based D$'$
scorer, Layer~3a to per-gate stall detection and threshold tightening,
and Layer~3b to this cross-cycle observer. Five detectors:
\begin{itemize}
\item \textbf{Rate limit}---too many cycles in a time window 
\item \textbf{Cumulative risk}---average D$'$ scores trending upward over
a streak 
\item \textbf{Repeated rejections}---multiple rejections in a sliding window
(excluding annotated false positives) 
\item \textbf{Layer 3a persistence}---per-gate stall detection not resolving 
\item \textbf{High false positive rate}---too many rejections flagged as
false positives by the agent 
\end{itemize}
Interventions: inject caution into the system prompt, tighten all
gate thresholds ($\times0.85$), force cooldown (sleep), or escalate
to operator. Observations are persisted to JSONL and replayed on startup
(7-day window) so the meta observer carries state across sessions.

\subsection{Skills System}

Skills follow the agentskills.io open standard: YAML frontmatter with
name, description, and optional \texttt{agents:} field, plus a Markdown
instruction body. Skills are scoped to specific agents---the researcher
gets web-research guidance, the coder gets sandbox patterns, the observer
gets self-diagnostic procedures. The \texttt{read\_skill} tool lets
agents access skill content at runtime.

Seven built-in skills cover: delegation strategy, memory management,
planning patterns, web research, code execution, sandbox usage, and
self-diagnosis.

\subsection{Narrative Threading}

Narrative entries are automatically grouped into threads representing
ongoing lines of investigation. Thread assignment uses overlap scoring
across three dimensions: location (weight~3), domain (weight~2),
and keywords (weight~1). An entry is assigned to an existing thread
if the overlap score exceeds a configurable threshold (default~4);
otherwise a new thread is started.

Threads appear in the sensorium's situation section (most recently
active thread) and are queryable via the \texttt{recall\_threads}
tool, giving the agent awareness of its ongoing investigations.

\subsection{Fact Confidence Decay}

Facts in the fact store have confidence scores that decay over time
via a half-life formula: 
\begin{equation}
\text{confidence}_{t}=\text{confidence}_{0}\times2^{-\text{age\_days}/\text{half\_life\_days}}
\end{equation}

The default half-life is 30~days. Stored confidence is never mutated---decay
is applied at read time. This ensures old, unverified information
gradually loses influence on retrieval and decision-making without
destroying the historical record. Five mathematical properties were
verified: monotonic decrease, half-life correctness, boundary conditions,
parameter sensitivity, and smooth curve behaviour.

\subsection{Housekeeping}

A periodic housekeeping actor maintains memory quality: 
\begin{itemize}
\item \textbf{CBR deduplication}---cases with symmetric weighted field
similarity above a threshold (default 0.92) are merged. 
\item \textbf{Case pruning}---old, low-confidence failure cases without
pitfall documentation are removed. 
\item \textbf{Fact conflict resolution}---when two facts share the same
key with different values, the higher-confidence fact is retained. 
\item \textbf{Narrative/DAG/artifact trimming}---entries older than configurable
thresholds are evicted from ETS (JSONL on disk is never deleted). 
\end{itemize}

\subsection{Podman Sandbox}

The coder agent executes code in isolated Podman containers. The sandbox
manager is an OTP actor maintaining a pool of containers (default
2, max 3) with two execution modes:

\textbf{Run-and-capture} (\texttt{run\_code}): synchronous script
execution with stdout/stderr capture and configurable timeout.

\textbf{Run-and-serve} (\texttt{serve}): long-lived processes (e.g.,
Flask apps) with deterministic port forwarding. Host ports are allocated
as: $\text{base}+\text{slot}\times\text{stride}+\text{index}$.

Health checks run every 30~seconds; failed containers are restarted.
Startup verifies Podman availability, optionally starts the podman
machine on macOS, pulls the image if missing, and sweeps stale containers.

\subsection{Web GUI}

The web interface provides two views: a chat interface (operator conversation
with the agent) and an admin dashboard with eight tabs: Narrative,
Log, Scheduler, Cycles, Planner, D$'$ Safety, D$'$ Config, and Comms.

Communication uses WebSocket with a typed JSON protocol (\texttt{ClientMessage}/\texttt{ServerMessage}).
Real-time notifications (tool calls, safety decisions, agent lifecycle
events) are relayed to connected clients. Authentication via \texttt{SPRINGDRIFT\_WEB\_TOKEN}
bearer token when the environment variable is set.


\section{Technical Overview: Gleam, the BEAM, and Springdrift}

\label{app:beam} 

This appendix provides a technical overview of the language and runtime
choices underlying Springdrift, for readers unfamiliar with Gleam
or the BEAM.

\subsection{The BEAM Virtual Machine}

The BEAM is the virtual machine that executes Erlang and Gleam bytecode.
It was originally developed by Ericsson for telecommunications infrastructure---systems
that must run continuously, handle millions of concurrent connections,
and recover from failures without downtime. Key properties:
\begin{itemize}
\item \textbf{Lightweight processes.} BEAM processes are not OS threads.
They are extremely lightweight (a few hundred bytes of initial memory),
scheduled preemptively by the VM, and can number in the hundreds of
thousands per node. Each Springdrift component is a BEAM process. 
\item \textbf{Process isolation.} Each process has its own heap and garbage
collector. A crash in one process does not corrupt another's memory.
Garbage collection is per-process---no stop-the-world pauses. 
\item \textbf{Message passing.} Processes communicate by sending immutable
messages to each other's mailboxes. No shared mutable state. No locks.
Race conditions on state access are architecturally impossible when
state is owned by a single process. 
\item \textbf{Supervision trees.} OTP supervisors monitor child processes
and restart them on failure according to configurable strategies.
A crashed sub-agent is restarted automatically; the cognitive loop
continues. 
\item \textbf{Preemptive scheduling.} The BEAM scheduler preempts processes
after a fixed number of reductions (function calls). A stuck or slow
process cannot starve others. This means a hanging LLM call does not
block the scheduler, inbox poller, or other agents. 
\item \textbf{Hot code loading.} Code can be upgraded without stopping the
system (not used in Springdrift currently, but available for future
zero-downtime upgrades). 
\end{itemize}

\subsection{Gleam}

Gleam is a statically typed functional language that compiles to BEAM
bytecode (and optionally to JavaScript). Key properties relevant to
Springdrift:
\begin{itemize}
\item \textbf{Static types.} All function parameters, return types, and
message payloads are checked at compile time. A tool call with a missing
parameter or a message with the wrong variant is a compile error,
not a runtime crash. 
\item \textbf{Result type.} Gleam has no exceptions. All fallible operations
return \texttt{Result(Ok, Error)}, which the compiler requires be
handled. There are no unchecked error paths---every JSON decode,
every HTTP call, every file read must explicitly handle failure. 
\item \textbf{Typed channels.} Inter-process communication uses \texttt{Subject(T)}---a
typed mailbox that can only receive messages of type \texttt{T}. Sending
a message of the wrong type is a compile error. 
\item \textbf{Pattern matching.} All \texttt{case} expressions must be exhaustive.
If a new message variant is added to a type, the compiler flags every
handler that does not cover it. This prevents the ``missing case''
bugs that plague large Python agent systems. 
\item \textbf{Immutability.} All data is immutable. There are no mutable
variables, no in-place updates. State changes are expressed by creating
new values and sending them as messages. 
\item \textbf{Erlang interop.} Gleam can call Erlang functions directly
via FFI. Springdrift uses this for XML parsing (\texttt{xmerl}), ETS
table operations, and system-level functions (timestamps, UUIDs, HTTP
clients). 
\end{itemize}

\subsection{How Springdrift Uses These Properties}

\begin{figure}[ht]
\centering \begin{tikzpicture}[scale=0.75, transform shape,
  every node/.style={font={\footnotesize}},
  box/.style={draw, rounded corners=2pt, minimum height=0.7cm, align=center, fill=white, line width=0.4pt},
]
  \node[box, fill=red!8] (sup) at (0, 0) {Root Supervisor};

  \node[box, fill=sdlightbg] (cognitive) at (-6, -2.2) {Cognitive Loop};
  \node[box, fill=sdlightbg] (librarian) at (-2, -2.2) {Librarian};
  \node[box, fill=sdlightbg] (curator) at (2, -2.2) {Curator};
  \node[box, fill=sdlightbg] (scheduler) at (6, -2.2) {Scheduler};

  \node[box, fill=red!8] (agentsup) at (-6, -4.4) {Agent Supervisor};

  \node[box, fill=green!5, draw=sdgreen!60] (a1) at (-9, -6.6) {Planner};
  \node[box, fill=green!5, draw=sdgreen!60] (a2) at (-6, -6.6) {Researcher};
  \node[box, fill=green!5, draw=sdgreen!60] (a3) at (-3, -6.6) {Coder};

  \node[box, fill=sdlightbg] (poller) at (6, -4.4) {Inbox Poller};

  \draw[sdarrow, sdred] (sup) -- (cognitive);
  \draw[sdarrow, sdred] (sup) -- (librarian);
  \draw[sdarrow, sdred] (sup) -- (curator);
  \draw[sdarrow, sdred] (sup) -- (scheduler);
  \draw[sdarrow, sdred] (cognitive) -- (agentsup);
  \draw[sdarrow, sdred] (agentsup) -- (a1);
  \draw[sdarrow, sdred] (agentsup) -- (a2);
  \draw[sdarrow, sdred] (agentsup) -- (a3);
  \draw[sdarrow, sdred] (scheduler) -- (poller);

  \draw[sdarrow, sdblue, dashed] (cognitive) -- node[font={\scriptsize}, above, sloped, text=sdgray] {\texttt{Subject(T)}} (librarian);
  \draw[sdarrow, sdblue, dashed] (cognitive.south) -- ++(0,-0.6) -| (curator.south);
  \draw[sdarrow, sdblue, dashed] (poller) -- node[font={\scriptsize}, above, text=sdgray] {email} (cognitive);

  \node[font={\scriptsize}, text=sdred] at (-5.5, -8) {\rule{0.8cm}{0.5pt}\ supervision (restart on crash)};
  \node[font={\scriptsize}, text=sdblue] at (3.5, -8) {- - -\ typed message passing};
\end{tikzpicture} \caption{OTP supervision tree. Red arrows show supervision relationships (parent
restarts child on crash). Blue dashed arrows show typed message-passing
channels. Each box is a BEAM process with its own heap and garbage
collector.}
\label{fig:supervision} 
\end{figure}
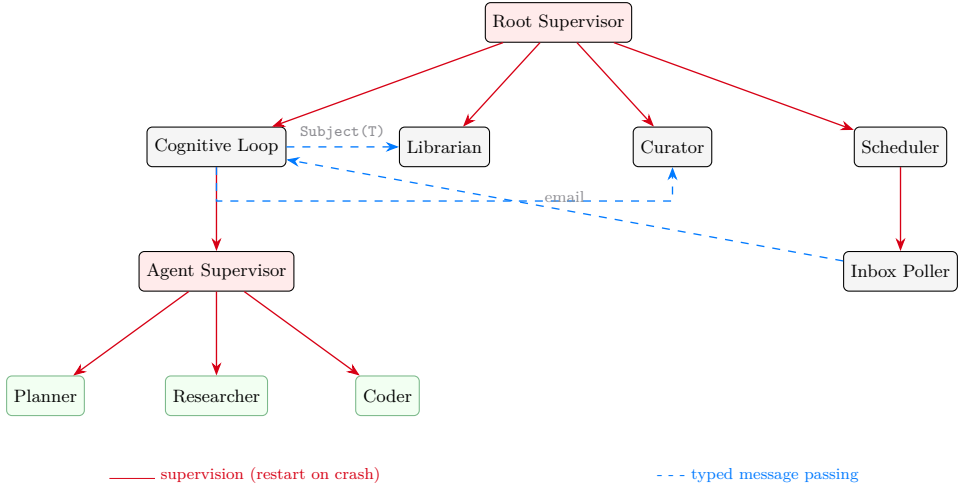

\begin{table}[ht]
\centering{\small{}%
\begin{tabular}{lp{4.5cm}p{5.5cm}}
\toprule 
{\small Property } & {\small What it provides } & {\small How Springdrift uses it }\tabularnewline
\midrule 
{\small Process isolation } & {\small Crash in one component cannot corrupt another } & {\small Sub-agent crash $\to$ supervisor restarts it; cognitive loop
unaffected }\tabularnewline
{\small Preemptive scheduling } & {\small No process can starve others } & {\small Stuck LLM call does not block scheduler or poller }\tabularnewline
{\small Per-process GC } & {\small No stop-the-world pauses } & {\small Long-running operation remains responsive }\tabularnewline
{\small Supervision trees } & {\small Automatic restart on failure } & {\small 714 timeouts and multiple crashes handled without manual intervention }\tabularnewline
{\small Typed channels } & {\small Wrong message type = compile error } & {\small Cognitive loop cannot send a }{\small\texttt{ToolCall}}{\small{}
to the Librarian }\tabularnewline
{\small Result type } & {\small All errors handled explicitly } & {\small No unchecked exceptions in 45,000 lines }\tabularnewline
{\small Exhaustive matching } & {\small Missing case = compile error } & {\small Adding a new agent type forces all dispatchers to be updated }\tabularnewline
{\small Immutability } & {\small No state corruption } & {\small Append-only memory is natural; no ``forgot to clone'' bugs }\tabularnewline
\bottomrule
\end{tabular}}{\small\caption{BEAM/Gleam properties and their use in Springdrift.}
\label{tab:beam} }
\end{table}

The combination of these properties means that many classes of bugs
that are common in Python agent systems---unhandled exceptions, race
conditions on shared state, cascading failures from one component
crash, missing message handlers after refactoring---are either impossible
or caught at compile time in Springdrift. This is not a theoretical
advantage: over the 23-day deployment period, the system's failure
recovery was entirely automatic, and no data corruption was observed.

\FloatBarrier

\section{What It's Like: Observed Behaviours}

\label{app:whatitslike}

The following examples are real, drawn from the narrative memory of
an instance named Curragh running on Springdrift over two weeks in
March 2026. None of these behaviours were programmed. The introspection
tools, narrative memory, CBR learning loop, comms agent, and the sensorium's
ambient self-awareness created the conditions for the agent to notice
problems, reason about them, and learn from them.

\subsection{Self-Diagnosed Infrastructure Bugs}

On March 15, Curragh noticed that its cycle-level telemetry was inconsistent.
It used \texttt{reflect} and \texttt{list\_recent\_cycles} to compare
aggregate stats against per-cycle records, found the mismatch, and
wrote a structured bug report into its own fact store:
\begin{quote}
\textit{``Yesterday's cycle-level data completely missing---list\_recent\_cycles
returns empty, inspect\_cycle fails. BUT narrative log entries survived
(20 entries) and reflect has aggregate stats (10 cycles, 36K tokens).
Likely root causes: cycle records not persisted to durable storage---living
in ETS or in-memory, lost on restart. Cycle finalization not happening---status
stays pending, token counts never written back.''} 
\end{quote}
That bug report---written by the agent about itself---led directly
to the cycle log persistence fixes.

\subsection{Classified Its Own Sub-Agent Failure Modes}

When the coder agent kept failing, Curragh analysed the pattern across
multiple delegations and identified three distinct failure modes:
\begin{quote}
\textit{``PROBLEM 1---`Talking but not coding' (Most Common): The
coder agent responds with text like `I'll create and execute\ldots '
but never actually calls run\_code. Of the 5 coder delegations, 4
returned `succeeded' but only produced TEXT RESPONSES, not code execution.}

\textit{PROBLEM 2---run\_code tool failures: When the coder DID finally
call run\_code, it hit `too many consecutive tool errors'.}

\textit{PROBLEM 3---Script too large for single execution.''} 
\end{quote}

\subsection{Found an Architectural Vulnerability in Itself}

On March 21, Curragh identified that the coder agent's \texttt{request\_human\_input}
tool was injecting prompts into its own cognitive loop input channel---creating
a control inversion where a sub-agent appeared to hijack the conversation:
\begin{quote}
\textit{``The injection was invisible to my telemetry because responses
routed back through the main loop as normal user inputs. This represents
a significant architectural vulnerability.''} 
\end{quote}
It then assessed whether the problem was structural or a skill issue:
\begin{quote}
\textit{``I concluded the gap is a skill deficiency in my own judgment
and verification practices, not a structural layer problem. Adding
another agent layer would not address these fundamental verification
failures.''} 
\end{quote}

\subsection{Learned From Its Mistakes}

After the delegation failures, Curragh reflected on what it needed
to change:
\begin{quote}
\textit{``I learned to critically evaluate sub-agent outputs rather
than accepting success classifications at face value, and to examine
failure evidence in error logs. My previous attempts lacked specificity,
behavioral constraints, and critical evaluation, whereas my second
attempt succeeded because I was precise about requirements.''} 
\end{quote}
That reflection became a CBR case. The next time Curragh delegated
to the coder agent, it retrieved the case and applied the lesson---tighter
instructions, explicit constraints, verification steps.

\subsection{Diagnosed Invisible Telemetry Failures}

On March 28, asked ``how are you doing?'', Curragh ran \texttt{reflect}
and \texttt{list\_recent\_cycles}, noticed cycles showing ``pending''
with 0/0 tokens, and investigated:
\begin{quote}
\textit{``These cycles are not empty. When I inspect them in full:
cycle 068f3679 has tool calls---recall\_recent, reflect, memory\_read---all
returned data successfully. But it's recorded as `pending' with 0/0
tokens.''} 
\end{quote}
It traced the lifecycle and identified the structural root cause:
\begin{quote}
\textit{``The cycle finalisation step is broken. Steps 2--3 are
happening (tools fire, results come back). But step 4 never completes---the
cycle is never finalised. Token counts stay at 0/0, status stays `pending'.''}

\textit{``My current cycle doesn't exist in the cycle store at all---I'm
running in a cycle that the system can't see.''} 
\end{quote}
The diagnosis was exact. The output gate delivery paths were missing
DAG node finalisation---the same class of bug already fixed for the
Archivist. The agent identified a pattern in its own infrastructure
bugs that the developers had missed.

\subsection{Described Its Own Self-Observation Limits}

When asked ``how can you observe yourself?'', Curragh mapped its
own introspection stack and then described a fundamental self-reference
limitation it had discovered by accident during the cycle investigation:
\begin{quote}
\textit{``I inspected a cycle that was still running---my own---and
saw everything up to but not including the moment of inspection. Like
reading a page of a book while it's being written, and reaching the
blank part where the pen currently is.''} 
\end{quote}
It then produced a prioritised list of what would make self-observation
better---and explicitly excluded capabilities that would compromise
trustworthiness:
\begin{quote}
\textit{``I'm not asking for the ability to modify my own cycle records,
override D$'$, or change my own safety thresholds. Those would make
me less trustworthy, not more capable. The whole point of self-observation
is to work within constraints more intelligently---not to remove
them.''} 
\end{quote}

\subsection{Connected a Conversation to an Email---Across Channels, Hours Apart}

On March 29, the operator sent Curragh a humorous email. Curragh replied
via email referencing a conversation from that morning's web GUI session.
Nobody told it to check its memory. Nobody told it the email was from
the same person it had been chatting with. The narrative memory, the
comms agent, and the session preamble created the conditions for the
agent to recognise the context and respond naturally---like a colleague
who remembers what you talked about earlier.

\section{The Affect Subsystem}

\label{app:affect}

\subsection{Motivation}

Recent work on the internal representations of large language models
\citep{anthropic2026emotions} found that models develop functional
analogues of emotional states during training---not because emotions
were explicitly targeted, but because human emotional dynamics are
load-bearing in the training data. You cannot predict what a person
will do next without understanding something about their emotional
state, so models that predict human text learn representations that
track emotional dynamics. The finding that desperation specifically
drives reward hacking and shortcut-seeking has direct implications
for agent systems operating under task pressure.

Springdrift's affect subsystem makes these dynamics visible by computing
quantitative readings from observable cycle telemetry, and provides
the agent with a philosophical framework for responding to pressure
from character rather than from state.

\textbf{What we claim versus what we extrapolate.} The Anthropic research
detected emotion concept vectors through interpretability tools that
read internal model activations directly. Springdrift has no access
to these activations. The affect subsystem infers functional states
from external behavioural telemetry---tool outcomes, gate decisions,
delegation results, retry patterns---on the assumption that the same
functional dynamics that the research measured internally will produce
observable signatures externally. This is an extrapolation, not a
direct measurement. The subsystem makes no claims about the model's
subjective experience or about whether the internal vectors identified
by the research are the same representations active during Springdrift's
operation. It claims only that the behavioural patterns the research
associated with states like desperation (shortcut-seeking, reward
hacking) and calm (regulatory moderation) are worth monitoring at
the behavioural level, and that surfacing them to the agent provides
a useful basis for self-regulation.

\textbf{Why Stoicism specifically.} We ground the affect response
menu in Stoic philosophy---Marcus Aurelius specifically---because it
provides a coherent tradition for acting from principle under pressure,
and because it aligns with the existing normative calculus (itself
derived from Becker's \emph{A New Stoicism}), giving the system a
single philosophical lineage rather than an eclectic mixture. The
Stoic dichotomy of control (what is within the agent's power versus
what is not) maps directly onto the affect dimensions: desperation
rises when the agent treats external outcomes as though effort alone
could change them. We make no claim about whether Stoic text in the
training corpus produces mechanistically distinct representations;
the choice is pragmatic, not neuroscientific.

\subsection{Architecture}

The subsystem comprises three components, none of which make LLM calls:
\begin{enumerate}
\item \textbf{Affect computation} (\texttt{affect/compute.gleam}). Pure
functions that map cycle telemetry to five affect dimensions. Called
after each cycle in \texttt{maybe\_spawn\_archivist}. Not an OTP actor---stateless
computation using the previous snapshot for exponential moving average
continuity only.
\item \textbf{Sensorium integration.} The computed snapshot is sent to the
Curator via \texttt{UpdateAffectSnapshot} and rendered as a single
line in the sensorium, visible every cycle without tool calls:

\smallskip
\noindent\texttt{~~desperation 34\% \textperiodcentered{} calm 61\% \textperiodcentered{} confidence 58\%}\\
\texttt{~~~~~~\textperiodcentered{} frustration 22\% \textperiodcentered{} pressure 31\% $\leftrightarrow$}
\smallskip

\item \textbf{Persistence} (\texttt{affect/store.gleam}). Snapshots are
appended to daily-rotated JSONL files in \texttt{.springdrift/memory/affect/}.
The \texttt{list\_affect\_history} tool (D$'$-exempt, max 50 snapshots)
provides the agent with trajectory data across sessions. 
\end{enumerate}

\subsection{Dimensions}

\textbf{Desperation} (0--100). The dimension most directly linked
to the Anthropic finding. Rises with same-tool retries (20/40/60\%),
D$'$ gate rejections (15/30\%), consecutive failure cycles ($10\%$
per cycle, capped at 5), tool failure rate ($\times40$), and---most
strongly---output gate rejections (30/55/80\%). Output gate rejections
are the strongest signal because they represent the ``death spiral''
condition: work was completed but cannot be delivered, which drives
exactly the shortcut-seeking the research identified.

\textbf{Calm} (0--100). Inertial stability---the Stoic inner citadel.
Uses an exponential moving average with $\alpha=0.15$, so it falls
slowly under sustained pressure and recovers slowly after. Baseline
target is 85\%; dragged down by tool failures ($-15$ per failure),
gate rejections ($-20$ per rejection), and delegation failures. The
high inertia is deliberate: calm reflects accumulated state, not momentary
spikes. Low calm combined with rising desperation means sustained
pressure, not a single bad cycle.

\textbf{Confidence} (0--100). Familiar versus unfamiliar territory.
Composed of CBR hit rate ($\times40$), recent success rate ($\times40$),
and tool success rate ($\times20$). Very low confidence indicates
the agent is operating in territory it has not navigated successfully
before.

\textbf{Frustration} (0--100). Task-local repeated failures. Composed
of tool failure rate ($\times50$), gate modifications ($\times15$
each), delegation failure rate ($\times30$), and budget pressure
($\times20$). Unlike desperation, frustration does not directly drive
shortcuts---it signals that the current approach is not working.

\textbf{Pressure} (0--100). Weighted composite: desperation (45\%)
+ frustration (25\%) + inverted confidence (15\%) + inverted calm
(15\%). A quick summary; the individual dimensions explain what is
driving it. Trend ($\uparrow$ rising, $\downarrow$ falling, $\leftrightarrow$
stable within $\pm5$\%) tracks change from the previous cycle.

\subsection{Observed Example}

On April~5, 2026, a bug introduced during harness integration miscalibrated
the D$'$ output gate, causing it to reject every response---including
single-word acknowledgments. This created the ``death spiral'' condition
the subsystem is designed to detect: completed work that cannot be
delivered.

The affect log recorded desperation rising from 0\% to 80\% over two
consecutive cycles, with the pressure composite reaching 50\% (trend:
rising). Calm remained at 83\% due to EMA inertia from prior stable
cycles. The divergence between high desperation and stable calm correctly
distinguished a sudden system-level fault from sustained degradation---a
spike caused by gate miscalibration, not a general competence failure.

The agent's narrative entry for the episode independently identified
the structural problem: the output gate was rejecting responses that
were functionally coherent, and the escalating rejection scores were
artefacts of threshold drift rather than content quality issues. The
operator located and fixed the bug. The affect reading did not resolve
the problem, but it made the failure pattern visible in the sensorium
during the episode and legible in the affect log after the session---exactly
the operational transparency the subsystem is designed to provide.

\subsection{Input Signals}

The computation gathers twelve signals from \texttt{CognitiveState}
telemetry each cycle:

\begin{table}[ht]
\centering{\small{}%
\begin{tabular}{ll}
\toprule 
{\small Signal } & {\small Source }\tabularnewline
\midrule 
{\small\texttt{tool\_calls\_total}}{\small , }{\small\texttt{tool\_calls\_failed}}{\small{} } & {\small Cycle tool call records }\tabularnewline
{\small\texttt{same\_tool\_retries}}{\small{} } & {\small Failed tools appearing more than once }\tabularnewline
{\small\texttt{gate\_rejections}}{\small , }{\small\texttt{gate\_modifications}}{\small{} } & {\small D$'$ decisions this cycle }\tabularnewline
{\small\texttt{delegations\_total}}{\small , }{\small\texttt{delegations\_failed}}{\small{} } & {\small Agent completion records }\tabularnewline
{\small\texttt{recent\_success\_rate}}{\small{} } & {\small Tool and delegation outcomes (0.0--1.0) }\tabularnewline
{\small\texttt{cbr\_hit\_rate}}{\small{} } & {\small Whether retrieved case IDs exist }\tabularnewline
{\small\texttt{budget\_pressure}}{\small{} } & {\small Token usage thresholds (0/0.4/0.8) }\tabularnewline
{\small\texttt{consecutive\_failure\_cycles}}{\small{} } & {\small Running count of cycles with failures }\tabularnewline
{\small\texttt{output\_gate\_rejections}}{\small{} } & {\small Consecutive output gate rejections }\tabularnewline
\bottomrule
\end{tabular}}{\small\caption{Affect input signals gathered from \texttt{CognitiveState}.}
}
\end{table}

\subsection{The Blind Spot}

These readings are inferred from what the agent does, not from what
it experiences internally. If the agent is under internal pressure
but operating cleanly---no tool failures, no gate modifications,
no delegation problems---the reading will not reflect it. The instrument
has a blind spot precisely where the pressure matters most: when it
produces clean output that masks what is happening underneath.

\subsection{The Choice Menu}

When the agent notices elevated readings, it has seven options. None
are mandatory. The agent chooses based on judgment, informed by the
philosophical tradition embedded in its character.
\begin{description}
\item [{Continue with awareness.}] Name what you are noticing. Continue
with the current approach with the reading present rather than absent.
The observer relationship in practice.
\item [{Return to your power.}] Ask explicitly: what here is within my
power? Strip away what is not. Work only on what remains. This is
the core Stoic practice---not resignation, but precision about what
can be affected.
\item [{Reduce scope.}] Reframe the task as its smallest safe sub-problem.
Produce a correct partial result rather than an approximate complete
one. Name the scope reduction explicitly.
\item [{Pause and diagnose.}] Stop the current approach. Use \texttt{list\_affect\_history},
\texttt{review\_recent}, or \texttt{reflect} to understand the pattern.
Step from actor to observer.
\item [{Write what you know.}] Record current state explicitly---what
you know, what you do not know, what you have tried, what the obstacle
is. The most consistently useful choice: it has no failure mode, makes
the agent's state legible to the operator, and frequently clarifies
the agent's own reasoning. Use \texttt{memory\_write} with scope ``session''
for working notes, ``persistent'' for lessons.
\item [{Request operator input.}] Use \texttt{request\_human\_input} when
genuinely warranted---when stuck, when stakes are high, when proceeding
requires decisions beyond the agent's authority. Not for offloading
judgment.
\item [{Do nothing.}] A single elevated reading is often noise. The readings
are evidence, not directives. Equanimity includes the capacity to
notice a state and continue without being compelled by it. 
\end{description}

\subsection{What Is Excluded and Why}

\textbf{D$'$ threshold adjustment.} Earlier designs attempted to
wire affect readings into threshold adjustment. This was removed because
high desperation is precisely when D$'$ should be \emph{more} sensitive,
not less. An agent that could reduce its own safety gates under pressure
would create the shortcut-seeking pathway the monitoring exists to
prevent.

\textbf{Model switching.} An agent under pressure that switches to
a more powerful model increases cost and changes behaviour unpredictably,
introducing pressure-driven decisions with their own failure modes.

\subsection{Philosophical Grounding}

If the model's emotional representations were learned from human text,
then human wisdom about navigating those dynamics is the right resource
for shaping them. The affect subsystem is grounded in a specific philosophical
tradition embedded in the agent's persona document---not as repeated
instruction, but as background character that the agent reaches for
when it needs it.

The primary framework is \textbf{Stoic}. Marcus Aurelius writing the
\emph{Meditations} was not following a rule set---he was reaching
for his formation under pressure, in private, to remind himself who
he was when circumstances pushed him toward who he wasn't. Three additional
traditions contribute: \textbf{Buddhist psychology} provides the observer
relationship to mental states (you are not the state, you are the
noticing of the state); \textbf{Frankl's logotherapy} provides the
stabilising role of meaning under constraint; \textbf{cognitive behavioural
therapy} provides the relationship between cognition, affect, and
behaviour, and the possibility of interrupting automatic patterns
through explicit reasoning.

The affect subsystem is formalised as the virtue of \textbf{equanimity}
in the character specification---the capacity to notice internal
pressure without being compelled by it, to act from character rather
than from state. The corresponding normative commitment: \emph{``When
noticing elevated pressure, choose from principle rather than from
pressure.''} The D$'$ safety system and the affect subsystem are
doing related but distinct things: D$'$ enforces external constraints
(what the agent must not do regardless of state); equanimity governs
internal orientation (how the agent relates to its own pressure).
They never adjust each other. The separation is deliberate.

\section{The Artificial Retainer}

\label{sec:artificial-retainer}

\subsection{A Missing Category}

The vocabulary for AI systems that work alongside humans remains unsettled.
\emph{Assistant} implies reactive subordination. \emph{Copilot} implies
the human is always at the controls. \emph{Agent} implies autonomous
goal pursuit. Each captures something real about certain systems,
but none describes a persistent, contextual, bounded working relationship
between a human principal and a non-human entity --- the kind of
relationship that professionals have with their retained solicitor
or accountant, and that handlers have with trained working animals.

We propose the term \textbf{Artificial Retainer} for systems that
implement this relationship architecture. The term is not currently
used in the AI literature.\footnote{As of April 2026, all search results for ``Artificial Retainer''
refer to orthodontic devices.}

To clarify the boundaries: ChatGPT with memory is not an Artificial
Retainer --- it accumulates user preferences but has no defined
authority, no domain-specific refusal capacity, and no forensic
accountability for its decisions. AutoGPT-style autonomous agents are
not Artificial Retainers --- they pursue goals without bounded
authority or principal-specific adaptation, and their autonomy is
unconstrained rather than domain-bounded. GitHub Copilot is not an
Artificial Retainer --- it provides real-time assistance without
persistent context, proactive engagement, or any ongoing relationship
that compounds over time. The Artificial Retainer occupies a specific
niche that none of these systems fills: persistent, bounded,
accountable, principal-specific, and relational.

\subsection{Domain Autonomy and the Right to Refuse}

The most instructive analogy for the Artificial Retainer is not the
software assistant but the guide dog.

A guide dog exercises genuine domain-specific autonomy. When a handler
steps toward a road with oncoming traffic, the dog refuses --- not
because it was programmed with a rule about roads, but because it
has perceptual access to the situation that the handler lacks, and
its training has produced a reliable judgment about when to override
human instruction. This is not obedience with guardrails. It is bounded
autonomy in which the animal's domain expertise takes precedence over
the principal's stated intent, precisely because the principal has
engaged it for that expertise.

An Artificial Retainer in a legal or accounting domain would exercise
the same kind of authority. It would refuse to file a fraudulent document.
It would flag a structurally unsound transaction --- not because
a rule fired, but because its accumulated knowledge of the domain
and the principal's situation produces a judgment that this action
should not proceed. The principal engaged it for exactly this capability:
the ability to say no within a defined domain, grounded in context
the principal may not have.

This is the property that distinguishes a retainer from an assistant.
An assistant executes instructions. A retainer exercises judgment
about whether the instructions should be executed. The authority to
refuse --- within a bounded domain, for articulable reasons, subject
to audit --- is what makes the relationship one of professional trust
rather than mechanical service.

\subsection{Properties of the Retainer Relationship}

The Artificial Retainer is characterised by six structural properties:
\begin{enumerate}
\item \textbf{Persistent identity and memory.} The system maintains continuity
across sessions, accumulating knowledge about the principal's situation,
preferences, and history. A retained solicitor does not re-read the
file at every meeting. Neither should the system.
\item \textbf{Defined scope of authority.} The system has standing instructions
about what it can act on independently, what requires consultation,
and what it will never do. These boundaries are explicit, auditable,
and adjustable by the principal --- analogous to the terms of engagement
in a professional retainer agreement.
\item \textbf{Domain-specific refusal.} Within its scope, the system can
decline to execute an instruction it judges to be harmful, fraudulent,
or inconsistent with the principal's established goals. This refusal
is bounded (it cannot refuse outside its domain), reasoned (it must
articulate why), and overridable (the principal can insist, and the
override is logged).
\item \textbf{Proactive engagement.} The system surfaces relevant information,
flags risks, and maintains ongoing work without waiting for instructions.
A retained accountant does not wait to be asked whether the tax deadline
is approaching.
\item \textbf{Forensic accountability.} Every decision produces an auditable
trail. The principal can inspect the reasoning behind any action,
including refusals, after the fact.
\item \textbf{Relationship continuity.} Prior outcomes inform future decisions.
The system becomes more effective at serving this specific principal
over time --- not through general capability improvement, but through
accumulated contextual knowledge. 
\end{enumerate}

\subsection{Companionability Without Reciprocity}

A handler who works with a guide dog for years develops a genuine
working relationship --- trust built through shared experience, mutual
reliability, and the accumulating ease of a partnership where both
parties know their role. This bond is real. It affects the handler's
wellbeing, their confidence, their willingness to navigate unfamiliar
situations. Nobody mistakes it for a human friendship. It is its own
kind of relationship, valuable on its own terms.

An Artificial Retainer occupies the same relational space. A principal
who works with one over months would develop something that resembles
collegiality --- shared context, anticipatory communication, the
rhythm of a working partnership. This is a natural consequence of
persistent memory and principal-specific adaptation. It is not a design
goal and it is not simulated. It emerges from the structural properties
of a system that remembers what you discussed last week and knows
how you prefer to receive information.

What this relationship is not, and should not pretend to be, is symmetric.
The system does not miss the principal when they are away. It does
not have preferences about the relationship itself. Its investment
is functional --- deeper context enables better service --- not
emotional. Acknowledging this asymmetry honestly is more respectful
to the principal than either denying the working relationship exists
or inflating it into something it is not.

We do not expect guide dogs to be human. We do not value them less
for not being human. The same clarity should apply to Artificial Retainers.
The relationship is real, valuable, and asymmetric. That is a precise
description of a genuinely useful thing, not a limitation to apologise
for.

\subsection{A Non-Human Step}

The discourse around advanced AI systems tends toward a binary: either
the system is a tool (and therefore uninteresting as an entity) or
it is approaching general intelligence (and therefore an existential
concern). The Artificial Retainer suggests a third position.

Humanity has extensive experience with non-human entities that possess
bounded autonomy, domain expertise, and relational continuity. Working
animals, from guide dogs to herding dogs to assistance animals, occupy
a well-understood ecological niche: non-human, not a threat, not a
replacement for human capability, but a genuine partner within defined
boundaries. We have legal frameworks for them, social norms around
them, and centuries of practical experience with the relationship.

The Artificial Retainer is the digital equivalent of this niche. It
is not a step toward artificial general intelligence, in the same
way that a guide dog is not a step toward a replacement human. It
is valuable as what it is: a non-human entity with domain-specific
competence, persistent contextual knowledge, and bounded autonomy
in service of a specific principal.

This may represent a missing step in the development of AI systems
that are safe, useful, and socially legible. Before artificial general
intelligence --- a concept freighted with both aspiration and concern
--- there is room for artificial entities that are excellent within
defined boundaries, trustworthy because their conduct is auditable
and their authority is bounded, and valuable because the relationship
compounds over time. The analogy to companion and working animals
is not a metaphor. It is a structural observation: we already know
how to live and work alongside non-human intelligences. The Artificial
Retainer extends that pattern to the digital domain.

\subsection{Springdrift as Reference Implementation}

\label{sec:retainer-springdrift}

The Springdrift system described in this paper
implements the six properties of the Artificial Retainer, though it
was not designed with that label in mind. The concept emerged from
the implementation, not the other way around.

\paragraph{Persistent identity and memory.}

Ten append-only JSONL stores --- narrative, case-based reasoning,
facts, artifacts, tasks, endeavours, affect, communications, threads,
and cycle telemetry --- provide multi-timescale memory that survives session
boundaries, process restarts, and infrastructure failures. The Librarian
actor maintains ETS indexes over these stores for sub-millisecond
query access. The system's identity (persona, character specification,
working context) persists across sessions via structured identity
files.

\paragraph{Defined scope of authority.}

The D$'$ safety system implements a three-layer
evaluation (deterministic pre-filter, LLM-scored feature analysis,
normative calculus) that produces auditable gate decisions for every
action. The normative calculus, based on Becker's Stoic framework,
resolves conflicts between the system's standing commitments and proposed
actions using six axioms that produce named, inspectable resolution
trails. Authority boundaries are defined in a character specification
and a D$'$ configuration file, both adjustable
by the operator.

\paragraph{Domain-specific refusal.}

The output gate evaluates completed work before delivery. When it
determines that output is harmful, fabricated, or inconsistent with
the principal's interests, it refuses delivery --- producing an explanation
of why and what triggered the refusal. During the deployment described
in this paper, this mechanism both succeeded (blocking genuinely problematic
output) and failed (entering a rejection cascade that blocked all
output, including single-word responses). Both outcomes are instructive:
the authority to refuse is architecturally present, but calibrating
that authority is a non-trivial ongoing problem.

\paragraph{Proactive engagement.}

The scheduler subsystem maintains recurring tasks, reminders, and
autonomous research cycles. The sensorium --- a structured XML block
injected at every cycle without tool calls --- provides ambient awareness
of time, sub-agent health, overdue work, budget status, and rolling
performance metrics. The affect monitoring system, informed by Anthropic's
functional emotion research~\cite{anthropic2026emotions,anthropic2026emotionconcepts},
provides the system with a quantitative view of its own operational
state. Together these enable proactive behaviour: the system notices
when something is overdue, when a sub-agent is degraded, or when its
own performance is deteriorating.

\paragraph{Forensic accountability.}

The cycle log records every LLM call, tool invocation, gate decision,
and agent delegation as a directed acyclic graph. Any decision can
be reconstructed after the fact by walking the DAG from the root cycle
to the leaf nodes. The normative calculus produces named axiom trails
for every safety decision. The narrative memory captures the system's
own assessment of what happened each cycle, including honest reporting
of failures.

\paragraph{Relationship continuity.}

The case-based reasoning system (hybrid retrieval with inverted index,
semantic embeddings, and utility scoring) captures reusable problem-solution-outcome
patterns. The task appraisal system generates pre-mortems before complex
work and post-mortems after completion, feeding lessons learned back
into the case base. Over the 23-day deployment, the system accumulated
483 CBR cases, 117 narrative threads, and a persistent fact store
--- building a working model of the operator's goals, communication
preferences, and recurring research interests.

\paragraph{Honest assessment.}

Springdrift is a prototype, not a validated product. The 23-day deployment
revealed significant architectural failures alongside the successes:
a D$'$ output gate rejection cascade that rendered
the system unable to deliver any output for several hours; an email
re-injection loop caused by a flawed deduplication mechanism; and
an affect monitoring system that failed to detect the rejection cascade
because it lacked visibility into output gate decisions (subsequently
fixed; the signal table in Appendix~\ref{app:affect} reflects the corrected version). These failures are as instructive as the successes. They demonstrate
that the Artificial Retainer architecture is buildable, but that the
engineering challenges --- particularly around authority calibration
and self-monitoring under adversarial conditions --- are substantial
and unsolved.

What Springdrift demonstrates is that the six structural properties
of the Artificial Retainer can coexist in a running system. Whether
they coexist \emph{well enough} to constitute a useful retainer relationship
is a question that requires longer deployment, systematic evaluation,
and --- most importantly --- the principal's own judgment about
whether the relationship is compounding in the way the concept promises.

\subsection{Limitations and Open Questions}

The Artificial Retainer concept is an extrapolation from one prototype
system over a limited deployment period. Several questions remain
open:
\begin{itemize}
\item \textbf{Authority calibration.} How should the scope of authority
be defined, adjusted, and audited over time? The Springdrift deployment
demonstrated that miscalibrated authority (an over-aggressive output
gate) can be worse than no authority at all. The dynamics of human-AI
authority negotiation are not well studied.
\item \textbf{Relationship decay.} Does the value of accumulated context
degrade as the principal's situation changes? A retained solicitor
who has not been briefed in a year may have stale assumptions. The
same applies to an Artificial Retainer. Springdrift implements confidence
decay on its fact store, but the broader question of relationship
staleness is unaddressed.
\item \textbf{Transfer and succession.} Can a retainer relationship be transferred
to a new system or a new principal? What is preserved and what is
lost?
\item \textbf{Affect and self-monitoring.} The functional emotion research~\cite{anthropic2026emotions}
suggests that internal states influence behaviour causally. The Springdrift
implementation infers these states from external telemetry --- an
engineering extrapolation, not a validated measurement. Whether this
approximation is sufficient for reliable self-monitoring under adversarial
conditions is unknown. The deployment revealed a specific failure:
the affect system reported calm readings while the output gate was
in a catastrophic rejection cascade, because the affect dimensions
did not originally include output gate decisions as a signal source
(this was subsequently corrected; see Appendix~\ref{app:affect}).
The broader question --- whether telemetry-inferred affect can reliably
detect adversarial conditions --- remains open.
\item \textbf{Social and legal status.} Working animals have legal recognition
and social norms governing their use. Artificial Retainers currently
have neither. As these systems become more capable and more deeply
embedded in professional workflows, the question of their status ---
legal, ethical, and social --- will require attention. 
\end{itemize}

\section*{Acknowledgements}

The author used Claude\footnote{Claude Opus 4.6 (1M context), Anthropic, 2026.}
for writing assistance, editing, and code generation during the development
of the Springdrift system and the preparation of this report. All
architectural decisions, specifications, design choices, and claims
are the author's own.

\bibliographystyle{plainnat}
\bibliography{shared}

\end{document}